\documentclass[10pt,table]{article}

\usepackage[preprint]{tmlr}
\usepackage{hyperref}
\usepackage{url}
\hypersetup{hidelinks}

\usepackage{latexsym}

\usepackage[T1]{fontenc}

\usepackage[utf8]{inputenc}

\usepackage{microtype}

\usepackage{amsmath, amsfonts, amssymb}
\usepackage{dsfont}
\usepackage{xspace}
\usepackage{tabularx}
\usepackage{graphicx}
\usepackage{wrapfig}
\usepackage{multicol}
\usepackage{multirow}
\usepackage{enumitem}
\usepackage{pifont}
\usepackage{booktabs}

\usepackage{graphicx}

\newtheorem{proposition}{Proposition}
\newtheorem{theorem}{Theorem}

\def\xv{\mathbf{x}}

\def\yv{\mathbf{y}}

\title{Verifier-Guided Self-Consistency: \\ Test-Time Reasoning with Imperfect Generators and Verifiers}

\author{
  \begin{tabular}{c}
    \large
    Ekaterina Fadeeva\textsuperscript{1} \quad
    Xueyan Li\textsuperscript{1,3} \quad
    Mrinmaya Sachan\textsuperscript{1} \quad
    Maxim Panov\textsuperscript{2}
    \\[0.8em]
    \textsuperscript{1}ETH Zürich
    \qquad
    \textsuperscript{2}MBZUAI
    \qquad
    \textsuperscript{3}Max Planck Institute for Intelligent Systems
  \end{tabular}
}

\begin{document}

\newcommand{\qwen}{Qwen2.5 Math 1.5B Instruct\xspace}
\newcommand{\gemma}{Gemma3 4B Instruct\xspace}
\newcommand{\llama}{Llama 3.2 3B Instruct\xspace}
\newcommand{\qwenprm}{Qwen2.5 Math PRM 7B\xspace}
\newcommand{\reprobe}{ReProbe\xspace}
\newcommand{\gpt}{GPT 4.1\xspace}
\newcommand{\pending}[1]{\textcolor{red}{#1}}
\newcommand{\todo}{\textcolor{red}{[TODO]}\xspace}
\newcommand\freefootnote[1]{%
  \let\thefootnote\relax%
  \footnotetext{#1}%
  \let\thefootnote\svthefootnote%
}

\newcommand{\multirowcell}[1]{\begin{tabular}[c]{@{}c@{}}#1\end{tabular}}
\newcommand{\tcell}[2][c]{\begin{tabular}[t]{@{}#1@{}}#2\end{tabular}}
\newcommand{\pmstd}[1]{{\tiny\,$\pm$\,#1}}

\definecolor{darkgreen}{RGB}{0,100,0}
\definecolor{darkred}{RGB}{139,0,0}
\definecolor{darkyellow}{RGB}{204,153,0}

\maketitle
\begin{abstract}
  Large language models can solve many multi-step reasoning problems, but their reasoning remains brittle: plausible intermediate steps often lead to incorrect answers. Improving reasoning at test time requires balancing two imperfect signals: the generator, which provides diversity, and the verifier, which guides correctness. Existing approaches largely favor one side of this tradeoff. Verifier-weighted voting relies on trajectories sampled from the generator and cannot steer generation toward correct ones. By contrast, beam-style verifier-guided search can steer generation, but may over-concentrate on narrow trajectory sets and amplify verifier errors.
  We propose Verifier-Guided Self-Consistency (VGSC), a stochastic decoding framework that combines step-level correctness and future-outcome estimates to guide reasoning during generation. VGSC repeatedly proposes candidate next steps from the base language model and uses both signals in an acceptance/resampling procedure, while maintaining multiple trajectories for final answer-level self-consistency. We provide a probabilistic interpretation of VGSC as approximate sampling from a verifier-tilted trajectory distribution. Across multiple reasoning benchmarks, generator models, and verifier families, VGSC achieves higher average accuracy than representative baselines.
  \freefootnote{
    Correspondence:
    \href{mailto:efadeeva@ethz.ch}{efadeeva@ethz.ch},
    \href{mailto:maxim.panov@mbzuai.ac.ae}{maxim.panov@mbzuai.ac.ae}.
  }
  \freefootnote{
    Our implementation is available at
    \url{https://github.com/cant-access-rediska0123/vgsc}
  .}
\end{abstract}


\section{Introduction}

  Large language models (LLMs) have shown strong performance on multi-step reasoning tasks, especially when prompted to produce chain-of-thought (CoT) explanations~\citep{wei2023chainofthoughtpromptingelicitsreasoning} and when multiple reasoning trajectories are aggregated with self-consistency~\citep{wang2023selfconsistencyimproveschainthought}. However, their reasoning remains brittle: models can produce plausible intermediate steps that lead to incorrect answers~\citep{cobbe2021gsm8k, lightman2023letsverifystepstep}.

  A natural way to improve reliability is to augment the LLM generator with a learned \emph{verifier} trained to estimate whether a solution, or an intermediate reasoning step, is correct~\citep{lightman2023letsverifystepstep, wang-etal-2024-math}. At inference time, such verifiers can guide search or rerank candidates without modifying the underlying LLM~\citep{khalifa-etal-2023-grace, xie2023selfevaluationguidedbeamsearch, ni2026efficienttesttimescalingmultistep}.

  Existing inference-time uses of learned verifiers largely fall into two families. In \emph{post-hoc aggregation}, multiple complete reasoning trajectories are sampled from the base model, and a verifier is used afterward to rank them or weight their votes~\citep{cobbe2021gsm8k, wang2023selfconsistencyimproveschainthought, kang2025scalablebestofnselectionlarge}. This preserves diversity, but cannot help if correct trajectories are rarely generated. In contrast, verifier-guided decoding uses verifier information \emph{during decoding} to guide search over partial reasoning traces~\citep{yao2023treethoughtsdeliberateproblem, khalifa-etal-2023-grace, xie2023selfevaluationguidedbeamsearch, zhao2024probabilisticinferencelanguagemodels, yu-etal-2024-ovm}. Some methods rely on \emph{local} signals that assess the next step~\citep{khalifa-etal-2023-grace, xie2023selfevaluationguidedbeamsearch}, while others use more \emph{future-looking} signals that estimate whether a prefix is likely to admit a successful continuation~\citep{zhao2024probabilisticinferencelanguagemodels, yu-etal-2024-ovm, setlur2024rewardingprogressscalingautomated}.

%

\begin{figure*}[ht]
\centering
\includegraphics[width=\linewidth, trim=0cm 0cm 0cm 0cm, clip]{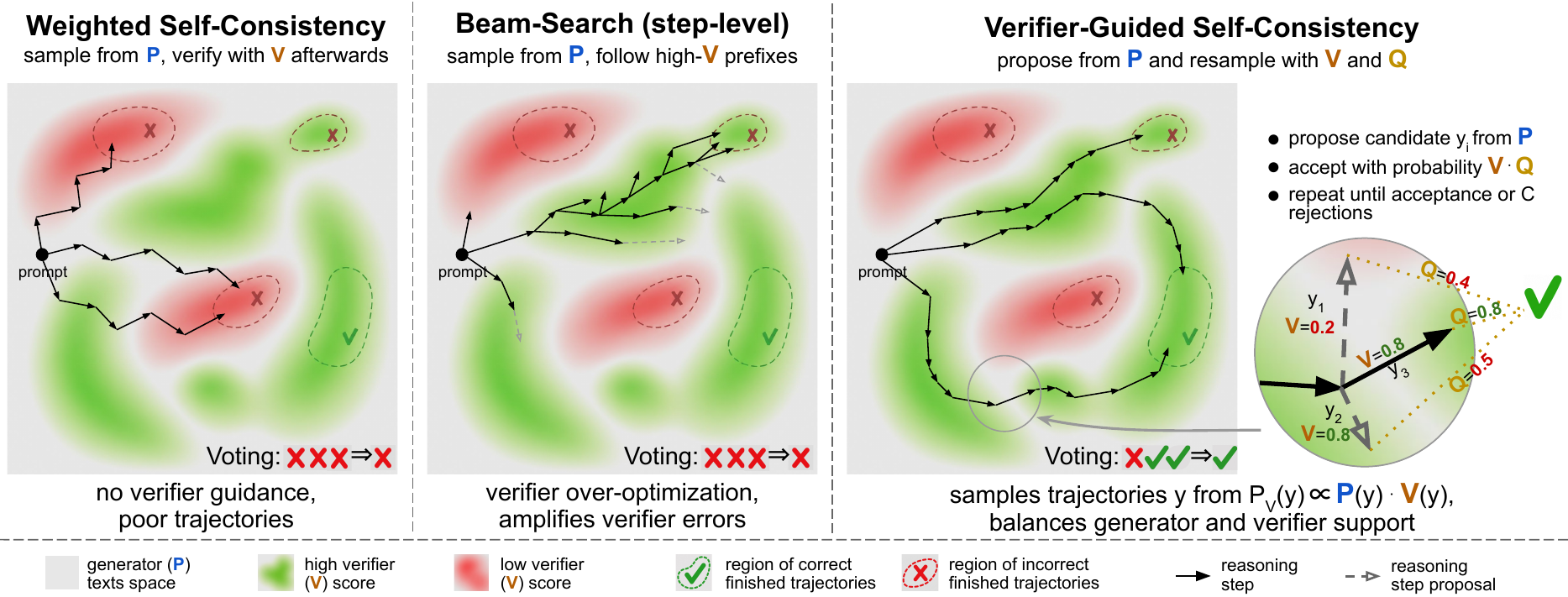}
\caption{
  Illustration of inference-time reasoning with imperfect generators and verifiers. 
  \textit{Left}: Weighted Self-Consistency samples complete trajectories from $P$ and applies verifier weighting after generation, so it cannot recover from poor samples. 
  \textit{Middle}: Beam Search uses verifier scores during decoding, but deterministic selection can concentrate search in a narrow high-$V$ region; with a noisy verifier, chains may converge to the same false-positive region. 
  \textit{Right}: VGSC samples candidate next steps from $P$ and uses bounded acceptance-resampling with $V$ and $Q$, balancing verifier guidance and trajectory diversity.
}
\label{fig:vgsc_illustration}
\end{figure*}

  These two families reflect a broader tension in test-time reasoning. The generator provides coverage and diversity over plausible trajectories, but may assign low probability to correct ones; the verifier provides guidance toward correctness, but is itself imperfect. Learned verifiers can be miscalibrated~\citep{park2025knowdontknowuncertainty}, generalize poorly across domains~\citep{prmlessons}, or be over-optimized when used too aggressively~\citep{khalaf2025inferencetimerewardhackinglarge}. Moreover, maximizing noisy scores inside beam search can introduce overestimation bias, so wider search may degrade reasoning performance rather than improve it~\citep{dalal2026testtimecomputehurtoverestimation}. Thus, post-hoc aggregation can fail when the generator rarely produces good trajectories, while beam-style verifier-guided search can fail when the verifier is too noisy for aggressive pruning.

  In this paper, we propose \textbf{Verifier-Guided Self-Consistency (VGSC)}, a decoding strategy for test-time reasoning with imperfect generators and verifiers. VGSC is motivated by the observation that neither generator likelihood nor verifier score is reliable enough to optimize in isolation. Instead of verifying only after complete generation or deterministically pruning to the highest-scoring partial chains, VGSC uses verifier information \emph{stochastically} during generation. At each reasoning step, it samples candidate next steps from the base LLM and resamples among them using two complementary signals: (i) a \emph{step verifier} estimating immediate correctness, and (ii) a \emph{future-outcome predictor} estimating whether the resulting prefix is likely to admit a successful continuation. Repeating this process yields multiple verifier-guided but diverse trajectories, whose final answers are aggregated by self-consistency. Our contributions are as follows:
  \begin{itemize}[leftmargin=15pt]
    \item We introduce VGSC, a stochastic verifier-guided decoding algorithm that combines step-level correctness and future-outcome prediction through bounded rejection-resampling, while maintaining multiple trajectories for answer-level self-consistency.
    \item We derive the exact local conditional of a verifier-tilted trajectory distribution and analyze the approximation error introduced by bounded rejection-resampling, establishing a finite-cap total variation bound under exact guidance.
    \item We empirically show that VGSC improves accuracy over representative baselines across benchmarks, LLMs, and verifier families, while reducing LLM-generated tokens.
  \end{itemize}


\section{Background and Motivation}
  We study test-time search for chain-of-thought reasoning with a generator model and verifier guidance. Given input problem $\xv$, let $\yv=(y_1,\dots,y_T)$ be a reasoning trajectory of $T$ steps. We denote prefixes by $\yv_{<t}=(y_1,\dots,y_{t-1})$ and $\yv_{\le t}=(y_1,\dots,y_t)$, suffixes by $\yv_{>t}=(y_{t+1},\dots,y_T)$, and prefix extension by $\yv_{<t}\circ y=(y_1,\dots,y_{t-1},y)$.

  The base language model induces a distribution over reasoning trajectories,
  \begin{equation}
    P(\yv \mid \xv) = \prod_{t=1}^T P(y_t \mid \xv, \yv_{<t}).
  \end{equation}

  We assume access to a verifier that estimates whether a reasoning trajectory is correct. Concretely, let $c \in \{0,1\}$ denote the event that trajectory $\yv$ is correct for problem $\xv$, and define
  \begin{equation}
    V(\yv \mid \xv) \approx \Pr(c = 1 \mid \xv, \yv).
  \end{equation}
  Assuming a stepwise verifier, we factorize it as
  \begin{equation}
    V(\yv \mid \xv) = \prod_{t=1}^T V(y_t \mid \xv, \yv_{<t}),
    \label{eq:verifier}
  \end{equation}
  where $V(y_t \mid \xv, \yv_{<t}) \in [0, 1]$ is the probability that step $y_t$ is correct given that the previous steps $\yv_{<t}$ are correct. This interpretation is aligned with standard step-verifier supervision, where training prefixes are truncated at the first incorrect step~\citep{lightman2023letsverifystepstep}.

\subsection{Inference-Time Reasoning with Verifiers}

  Existing inference-time methods use verifier information in two main ways. The first is post-hoc aggregation: multiple complete reasoning trajectories are sampled from the base model and combined only after generation, for example by majority vote as in self-consistency (SC; \citealp{wang2023selfconsistencyimproveschainthought}) or by verifier-weighted voting as in weighted self-consistency (WSC; \citealp{kang2025scalablebestofnselectionlarge}). These methods preserve diversity across reasoning paths, but the verifier does not guide generation itself, so they can fail when the generator rarely, or never, samples a correct trajectory.

  The second family uses verifier information during decoding, for example through beam-style or tree-based search over partial reasoning traces~\citep{xie2023selfevaluationguidedbeamsearch, yao2023treethoughtsdeliberateproblem}. Such methods can allocate more computation to promising prefixes, but when verifier scores are noisy, they risk over-concentrating on a narrow region of the search space. Recent work shows that maximizing imperfect inference-time scores can induce overestimation bias, so increasing search width need not monotonically improve reasoning performance and can even degrade it~\citep{dalal2026testtimecomputehurtoverestimation}.

  These families expose a natural tradeoff: post-hoc aggregation preserves diversity but cannot steer generation toward missing correct trajectories, whereas verifier-guided search uses feedback earlier but can amplify local verifier errors and collapse diversity through aggressive deterministic pruning. Our goal is to combine the strengths of both families: generating diverse reasoning trajectories while biasing computation toward prefixes likely to lead to correct solutions. This motivates \textbf{Verifier-Guided Self-Consistency (VGSC)}, which uses verifier guidance stochastically during decoding while still producing multiple independent trajectories for final self-consistency. Rather than optimizing any one imperfect signal in isolation, VGSC combines generator likelihood, immediate step correctness, and continuation value. Figure~\ref{fig:vgsc_illustration} compares VGSC with representative post-hoc aggregation and verifier-guided search approaches.


\section{Method: Verifier-Guided Self-Consistency}
\label{sec:method}

\subsection{Verifier-tilted Target Distribution}
  We begin by defining the trajectory distribution that VGSC aims to approximate. Given a base LLM generator $P(\yv \mid \xv)$ and a verifier $V(\yv \mid \xv)$, we define the verifier-tilted distribution
  \begin{equation}
    P^V_\alpha(\yv \mid \xv)
    :=
    \frac{P(\yv \mid \xv) V(\yv \mid \xv)^\alpha}
    {\sum_{\yv'} P(\yv' \mid \xv) V(\yv' \mid \xv)^\alpha}.
    \label{eq:tilted_distribution}
  \end{equation}
  Here, $\alpha \ge 0$ controls the strength of verifier guidance: $\alpha=0$ recovers the base generator, $\alpha=1$ gives the standard verifier-tilted distribution~\citep{zhao2024probabilisticinferencelanguagemodels}, and larger $\alpha$ concentrates more mass on high-verifier trajectories. Thus, $P^V_\alpha$ favors trajectories that are both likely under the generator and highly rated by the verifier.

  This distribution has a simple probabilistic interpretation at $\alpha=1$. If the verifier is calibrated so that $V(\yv \mid \xv) = \Pr(c = 1 \mid \xv, \yv)$, then
  \begin{align}
    P(\yv \mid \xv) V(\yv \mid \xv) 
    = P(\yv \mid \xv)\Pr(c = 1 \mid \xv, \yv) = \Pr(\yv, c=1 \mid \xv),
  \end{align}
  and therefore $P^V_1(\yv \mid \xv) = \Pr(\yv \mid \xv, c=1)$.
  Thus, $P^V_1$ is the posterior over reasoning trajectories conditioned on correctness, while $P^V_\alpha$ is a tempered analogue. Weighted self-consistency method~\citep{kang2025scalablebestofnselectionlarge}, which samples from $P$ and weights resulting texts with verifier scores after generation, can be viewed as importance sampling for the answer marginal under $P^V_1$. VGSC instead uses verifier information during decoding to sample directly from $P^V_\alpha$.

\subsection{Exact Local Conditional}
  For a prefix $\yv_{<t}$, the verifier-tilted trajectory distribution induces the following exact next-step conditional:

  \begin{proposition}
    \label{prop:local_tilted_conditional}
    Given a prefix $\yv_{<t}$, define {\bf expected future score} as
    \begin{equation}
      \begin{aligned}
        Q_\alpha(\yv_{<t}\circ y, \xv)
        =
        \mathbb{E}_{\yv_{>t}\sim P(\cdot \mid \xv,\yv_{<t}\circ y)} \left[
          V(\yv_{>t} \mid \xv,\yv_{<t}\circ y)^\alpha
        \right].
      \end{aligned}
    \label{eq:q_equation}
    \end{equation}
    Then the next-step conditional of the verifier-tilted distribution $P^V_\alpha$ is
    \begin{align}
      P^V_\alpha(y_t = y \mid \xv,\yv_{<t})
      \propto P(y \mid \xv, \yv_{<t}) \cdot V (y \mid \xv, & \yv_{<t})^\alpha \cdot Q_\alpha(\yv_{<t}\circ y, \xv).
      \label{eq:verifier_tilted_conditional}
    \end{align}
  \end{proposition}
  The proof is provided in Appendix~\ref{app:proposition_proof}. Proposition~\ref{prop:local_tilted_conditional} shows that the ideal next-step distribution combines three signals: generator probability ($P$), immediate step correctness ($V$), and a continuation value ($Q$) measuring how promising the resulting prefix is. This combination is central to VGSC, making the method less dependent on any single imperfect signal.
  
  The main challenge is that $Q_\alpha(\yv_{<t}\circ y,\xv)$ is generally not available in closed form. Since the verifier is trained to approximate correctness probabilities, the future verifier factor $V(\yv_{>t} \mid \xv,\yv_{<t}\circ y)$ can be viewed as estimating whether a sampled continuation $\yv_{>t}$ completes the reasoning correctly from the current prefix. Thus, $Q_\alpha$ measures the expected verifier-rated success of continuations from $\yv_{<t}\circ y$. Estimating this expectation online with Monte Carlo rollouts at every decoding step would be prohibitively expensive. Instead, we use a learned future-outcome predictor $\widehat{Q}(\mathbf y_{\le t},\mathbf x)$ to estimate the probability that a continuation from the current prefix reaches the correct final answer.

  We train $\widehat{Q}$ using Monte Carlo continuation labels: for each prefix in the training data, we sample multiple continuations and use the fraction that reach the correct final answer as the target. At decoding time, we apply the same verifier-trust exponent $\alpha$ to both probability-valued signals:
  \begin{equation*}
      W_\alpha(y_t \mid \xv,\yv_{<t})
      = V(y_t \mid \xv,\yv_{<t})^\alpha
      \widehat{Q}(\yv_{\le t},\xv)^\alpha.
  \end{equation*}

\subsection{VGSC Decoding Algorithm}
\label{sec:vgsc_decoding_algo}
  Starting from an empty prefix, VGSC generates one trajectory by repeating a bounded rejection-resampling procedure. At step $t$, given prefix $\yv_{<t}$, define the local acceptance weight
  \begin{align}
    w_t(y) &:= W_\alpha(y \mid \xv, \yv_{<t}) = V(y \mid \xv, \yv_{<t})^\alpha \cdot \widehat{Q}(\yv_{<t} \circ y, \xv)^\alpha . \label{eq:local_acceptance_weight} 
  \end{align}
  Since $V$ and $\widehat{Q}$ take values in $[0,1]$, $w_t(y) \in [0,1]$. We then sequentially propose
  \begin{equation}
    y_t^{(1)}, y_t^{(2)}, \dots \overset{i.i.d.}{\sim} P(\cdot \mid \xv, \yv_{<t}),
  \end{equation}
  and accept proposal $y_t^{(j)}$ with probability $w_t(y_t^{(j)})$. If accepted, we set $\tilde y_t = y_t^{(j)}$ and stop sampling candidates for this step.

  To keep computation bounded, we allow at most $C$ proposals per step. If all are rejected, we \textit{fall back} to weighted selection among the cached proposals:
  \begin{equation}
    \Pr\!\left(\tilde y_t = y_t^{(j)} \mid \xv, \yv_{<t}\right) \propto \frac{w_t(y_t^{(j)})}{1-w_t(y_t^{(j)})}.
    \label{eq:vgsc_fallback_resampling}
  \end{equation}
  The selected step $\tilde y_t$ is appended to the prefix, and decoding continues until the trajectory is complete.

  This procedure is adaptive: high-weight proposals can be accepted early, reducing generated tokens relative to always sampling $C$ candidates. The cap $C$ bounds the worst-case per-step budget, while fallback ensures decoding can always proceed.

  We repeat this independently $N$ times to obtain guided trajectories $\yv^1,\dots,\yv^N$, extract the final answer $A(\yv^i)$ from each trajectory $\yv^i$, and predict by self-consistency:
  \begin{equation}
    \mathrm{VGSC}(\xv)=\arg\max_a\sum_{i=1}^N \mathds{1}\left( A(\yv^i) = a \right).
    \label{eq:vgsc}
  \end{equation}

\subsection{Theoretical Motivation}
  The following theorem quantifies the finite-cap bias introduced by bounded rejection-resampling.

  \begin{theorem}
    Fix a prefix $\yv_{<t}$, and define 
    \begin{equation}
      Z_t := \mathbb{E}_{Y\sim P(\cdot \mid \xv,\yv_{<t})}[w_t(Y)].
    \end{equation}
    Suppose $Z_t > 0$ and the practical local guidance weight is exact, so that
    \begin{equation}
      \begin{aligned}
        q_t(y) &:= P^V_\alpha(y_t=y \mid \xv,\yv_{<t}) = \frac{P(y \mid \xv,\yv_{<t})w_t(y)}{Z_t}.
      \end{aligned}
    \end{equation}
    Let $\mathcal{K}_{C,t,\alpha}(\cdot \mid \xv,\yv_{<t})$ be the next-step distribution of the bounded VGSC decoder in Section~\ref{sec:vgsc_decoding_algo}. Then the probability of reaching fallback is
    \begin{equation}
      \Pr(\mathrm{fallback} \mid \xv,\yv_{<t})
      =
      (1-Z_t)^C.
    \end{equation}
    Conditioned on not reaching fallback, the accepted proposal is distributed exactly as $q_t$. Moreover, assume that $0 < \epsilon \le w_t(y) \le 1 - \epsilon$. Then there exists a constant $B_t$, independent of $C$, such that
    \begin{equation}
      d_{\mathrm{TV}}\!\left(\mathcal{K}_{C,t,\alpha}(\cdot \mid \xv,\yv_{<t}),q_t(\cdot)\right)
      \le
      B_t \frac{(1-Z_t)^C}{C}.
    \end{equation}
    \label{theorem:vgsc}
  \end{theorem}
  Appendix~\ref{app:theorem_proof} provides the proof of Theorem~\ref{theorem:vgsc}.
  
\paragraph{Discussion.}
  Theorem~\ref{theorem:vgsc} shows that VGSC approximates the ideal verifier-tilted local conditional with only finite-cap bias. Before the proposal cap $C$ is reached, rejection-resampling is exact; approximation arises only when all $C$ proposals are rejected. This event has probability $(1-Z_t)^C \le \exp(-CZ_t)$, which decays exponentially in the proposal budget $C$ and the average acceptance mass $Z_t$. Thus, when candidates have non-negligible guidance weight, even a small cap makes fallback unlikely. With odds-weighted fallback, the resulting total variation error is of order $\exp(-CZ_t)/C$. Increasing $C$ therefore quickly reduces bias, while early acceptance preserves adaptive compute savings. Empirically, using only $C=5$ proposals per step is already sufficient to outperform strong baselines (see Section~\ref{sec:experiments}). We further measure in Appendix~\ref{app:fallback_statistics} that the fallback case is relatively infrequent: it occurs in about $46\%$ of decoding steps for $\alpha=1$ and remains below $71\%$ even under much stronger verifier guidance, $\alpha=20$.

  Appendix~\ref{app:proofs} further extends this one-step result to trajectory-level distributions and analyzes stability under imperfect verifier scores. Overall, Theorem~\ref{theorem:vgsc} shows that VGSC closely targets the verifier-tilted local conditional, and therefore inherits its balance of generator probability, step correctness, and continuation value. This balance reduces overconcentration on the failure modes of either the generator or the verifier.

  \begin{wrapfigure}{r}{0.4\textwidth}
    \centering
    \vspace{-1em}
    \includegraphics[width=\linewidth]{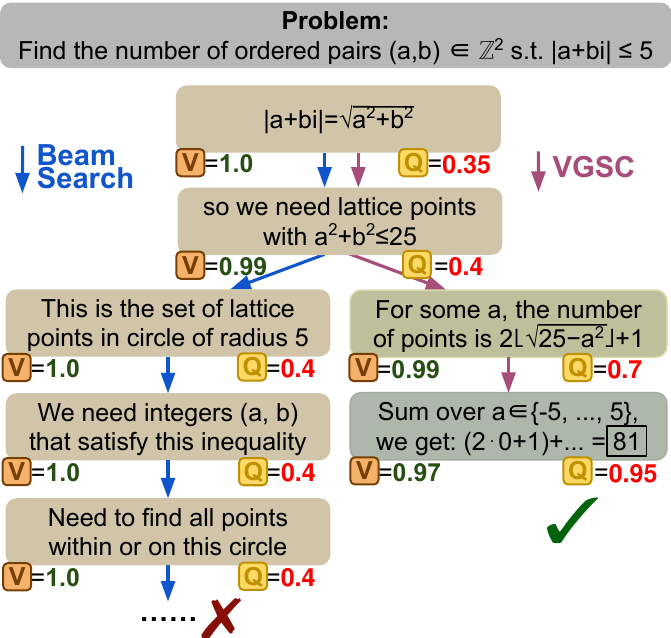}
    \caption{
      Simplified MATH example with \llama{} and a PRM verifier.
      The future-outcome score $Q$ favors prefixes likely to admit
      a correct continuation, steering VGSC toward the correct answer.
    }
    \label{fig:future_score_example}
    \vspace{-3em}
  \end{wrapfigure}

\subsection{Practical Motivation}
    
  The future-outcome term $Q$ is practically important because a locally correct step may still fail to advance the solution. For example, a model may restate the problem or enumerate constraints without introducing a useful strategy; such steps can receive high step-verifier scores, while their prefixes remain unlikely to admit a correct continuation. Figure~\ref{fig:future_score_example} illustrates this effect: step-level guidance alone can prefer locally plausible but stagnant continuations, whereas $Q$ favors prefixes more likely to lead to a correct solution. This motivates combining immediate correctness with future-outcome prediction during decoding.


\section{Experiments}
\label{sec:experiments}

\subsection{Experimental setup}

\paragraph{Datasets.}
  We evaluate on five reasoning benchmarks: MATH~\citep{hendrycksmath2021} and OlympiadBench~\citep{He2024OlympiadBenchAC} for mathematics, text-only ScienceQA~\citep{lu2022learn} for science, CommonsenseQA~\citep{talmor-etal-2019-commonsenseqa} for commonsense reasoning, and Knights\&Knaves~\citep{xie2024memorization} for logic. We use the test splits of MATH and OlympiadBench. All tasks require reasoning with extractable final answers. Table~\ref{tab:dataset_stats} summarizes dataset statistics; prompts, answer extraction, and examples are provided in Appendix~\ref{app:setup}.

  \begin{table*}[h]
\centering
\small
\caption{Summary statistics of the evaluation datasets. Mean number of reasoning steps is computed for \qwen using our rule-based step splitting procedure.}
\begin{tabular}{lcccc}
  \toprule
  Dataset & Domain & \# Samples & Mean \# Steps & Answer format \\
  \midrule
  MATH & Math & 500 & 8.4 & numeric \\
  OlympiadBench & Math & 675 & 4.6 & numeric \\
  ScienceQA (SciQA) & Science reasoning & 500 & 3.0 & multiple-choice \\
  CommonsenseQA (CsQA) & Commonsense reasoning & 500 & 9.9 & multiple-choice \\
  Knights\&Knaves (K\&K) & Logic & 100 & 10.5 & multiple-choice \\
  \bottomrule
\end{tabular}
\label{tab:dataset_stats}
\end{table*}

\paragraph{LLMs.}
  We use three instruction-tuned generators: \qwen~\citep{yang2024qwen25mathtechnicalreportmathematical} in the main text, with additional results for \llama~\citep{grattafiori2024llama} and \gemma~\citep{Kamath2025Gemma3T} in Appendix~\ref{app:additional_results}.

\paragraph{Verifiers.}
  Our primary step-level verifier $V$ is the frozen \qwenprm \citep{prmlessons,lightman2023letsverifystepstep}. For future-outcome prediction, we initialize a separate model from the same checkpoint and train $\widehat{Q}$ using LoRA adapters only in its last four of 28 Transformer layers \citep{hu2021loralowrankadaptationlarge}, yielding $46.1$M trainable parameters. All remaining parameters are frozen. For each training prefix derived from PRM800K, we sample 10 continuations and train $\widehat{Q}$ with binary cross-entropy against the fraction that reach the correct final answer. Because PRM800K is derived from the MATH training split, we treat MATH as in-domain for the verifier and the remaining benchmarks as out-of-domain.

  In Appendix~\ref{app:additional_results}, we also evaluate ReProbe~\citep{ni2026efficienttesttimescalingmultistep}, a lightweight ($\sim$20M parameters) verifier trained on hidden states of a frozen LLM. We train separate ReProbe models for step correctness and Monte Carlo continuation prediction. Appendix~\ref{app:verifier_training} provides more details on verifier training.

\paragraph{Evaluation Protocol.}
  We report exact-match accuracy of extracted final answers, averaged over $5$ seeds, and average generated tokens per example to measure the compute--accuracy tradeoff.

\subsection{Baselines}
  We categorize baseline methods according to how they use verifier information.

\paragraph{No-verifier baselines.}
  \textit{Single sample} returns one sampled trajectory. \textit{Self-consistency (SC)} samples $N$ trajectories and takes the majority-vote answer.

\paragraph{Post-hoc verifier aggregation.}
  \textit{Weighted self-consistency (WSC)} samples $N$ trajectories and weights each vote by the full-trajectory verifier score~\citep{kang2025scalablebestofnselectionlarge}. We also evaluate verifier-weighted \textit{DLE}~\citep{li_efficient_2026}, which uses a specialized sampling procedure to generate $N$ diverse trajectories and then applies weighted self-consistency over the resulting set.

\paragraph{Verifier-guided search baselines.}
  \textit{Beam search (BS)} uses verifier guidance during generation. We evaluate \textit{BS-Step}, which scores prefixes by the product of step scores $V$, and \textit{BS-Future}, which scores prefixes by $\widehat{Q}$. \textit{BS+SC} majority-votes over completed beam candidates. \textit{Twisted Sequential Monte Carlo (TSMC)} follows~\citet{zhao2024probabilisticinferencelanguagemodels}, using $\widehat{Q}$ for partial-generation twist estimates and $V(\yv \mid \xv)$ as the final chain potential.

\paragraph{VGSC.}
  VGSC jointly uses $V$ and $\widehat{Q}$ during decoding, as described in Section~\ref{sec:method}. We evaluate $\alpha \in \{1,5,10,20\}$, where larger $\alpha$ gives stronger verifier guidance.

  We use \texttt{ThinkBooster}~\citep{thinkbooster2026} implementations for all benchmarks and decoding strategies. Chain-of-thought traces are split into reasoning steps using a rule-based procedure; each step is capped at $2048$ tokens and each trajectory at $50$ steps. All experiments use temperature $1.0$. For BS, BS+SC, and VGSC, we use $N=5$ chains and $C=5$ candidates per chain; for SC and WSC, we evaluate $N\in\{5,25\}$; and for DLE and TSMC, we use $N=25$.
  Appendix~\ref{app:setup} provides full prompts, decoding settings, step-splitting details, and verifier-training hyperparameters. Appendix~\ref{app:baselines} describes the baseline methods, while Appendix~\ref{app:alternative_aggregations} reports additional baseline variants.

\subsection{Main Results}

\begin{table*}[h]
  \centering
  \small
  \caption{Accuracy $\uparrow$ across datasets and average generated tokens per sample for \qwen generator with \qwenprm verifier. Results are averaged over $5$ seeds. TFLOPs are estimated using Eq.~\eqref{eq:compute_formula};. The best result in each column is \textbf{bold} and the second-best result is \underline{underlined}.}
\resizebox{0.9\textwidth}{!}{
  \begin{tabular}{l|ccccc|ccc}
    \toprule
    Method & MATH & \multirowcell{Olympiad\\Bench} & SciQA & CsQA & K\&K & \multirowcell{Average\\Accuracy} & \multirowcell{Average\\Tokens} & \multirowcell{Estimated\\TFLOPs} \\
\midrule
 \rowcolor{gray!20}
 \multicolumn{9}{c}{\textit{No-verifier baselines}} \\
\midrule
Single sample & 70.6\pmstd{0.4} & 33.6\pmstd{0.4} & 56.0\pmstd{0.4} & 35.1\pmstd{0.7} & 44.2\pmstd{3.0} & 47.9\pmstd{0.6} & 587 & 2.1 \\
SC $N=5$ & 75.2\pmstd{0.6} & 35.4\pmstd{0.3} & 59.0\pmstd{0.7} & 40.9\pmstd{1.2} & 54.6\pmstd{1.4} & 53.0\pmstd{0.4} & 2{,}928 & 9.1 \\
SC $N=25$ & 76.5\pmstd{0.5} & 38.7\pmstd{0.3} & 63.4\pmstd{0.5} & 46.6\pmstd{0.3} & 56.0\pmstd{1.0} & 56.2\pmstd{0.3} & 14{,}778 & 44.6 \\
\midrule
 \rowcolor{gray!20}
 \multicolumn{9}{c}{\textit{Post-hoc verifier aggregation}} \\
\midrule
WSC $N=5$ & 78.0\pmstd{0.2} & 38.6\pmstd{0.3} & 61.5\pmstd{0.6} & 43.2\pmstd{0.5} & 51.4\pmstd{1.6} & 54.5\pmstd{0.4} & 2{,}964 & 52.0 \\
WSC $N=25$ & 80.5\pmstd{0.4} & 43.6\pmstd{0.2} & 67.0\pmstd{0.3} & 49.8\pmstd{0.7} & 56.2\pmstd{1.6} & 59.4\pmstd{0.4} & 14{,}791 & 253.1 \\
DLE & 78.6\pmstd{0.2} & 43.8\pmstd{0.2} & 67.5\pmstd{0.2} & 52.8\pmstd{0.4} & 60.4\pmstd{0.7} & 60.6\pmstd{0.2} & 14{,}655 & 248.8 \\
\midrule
 \rowcolor{gray!20}
 \multicolumn{9}{c}{\textit{Verifier-guided search baselines}} \\
\midrule
BS-Step & 77.4\pmstd{0.3} & 38.5\pmstd{0.4} & 44.8\pmstd{0.6} & 31.1\pmstd{0.5} & 38.0\pmstd{2.5} & 46.0\pmstd{0.5} & 19{,}837 & 338.9 \\
BS-Future & 80.1\pmstd{0.2}& 43.8\pmstd{0.4}& 64.8\pmstd{0.5}& 49.9\pmstd{0.4}& 55.0\pmstd{2.6}& 58.7\pmstd{0.5}& 12{,}337 & 211.4 \\
BS+SC-Step & 78.4\pmstd{0.1} & 41.1\pmstd{0.4} & 62.1\pmstd{0.7} & 46.7\pmstd{0.6} & 56.6\pmstd{1.2} & 57.0\pmstd{0.3} & 19{,}911 & 340.1 \\
BS+SC-Future & 79.0\pmstd{0.1}& 43.8\pmstd{0.4}& 66.6\pmstd{0.6}& 49.5\pmstd{0.7}& \textbf{64.5}\pmstd{2.3}& 60.7\pmstd{0.5}& 12{,}639 & 216.5 \\
TSMC & 78.7\pmstd{0.2}& 42.4\pmstd{0.4}& 65.2\pmstd{0.4}& 46.3\pmstd{0.8}& 55.0\pmstd{0.8}& 57.5\pmstd{0.3}& 15{,}045 & 287.7 \\
\midrule
 \rowcolor{gray!20}
 \multicolumn{9}{c}{\textit{Verifier-guided self-consistency (ours)}} \\
\midrule
VGSC $\alpha=1$ & 80.1\pmstd{0.3}& 43.3\pmstd{0.5}& 66.5\pmstd{0.5}& 51.1\pmstd{0.5}& 56.5\pmstd{2.1}& 59.5\pmstd{0.4}& 10{,}138 & 194.5 \\
VGSC $\alpha=5$ & \textbf{81.2}\pmstd{0.2}& 44.4\pmstd{0.3}& \underline{70.1}\pmstd{0.3}& 54.1\pmstd{0.5}& 59.5\pmstd{1.0}& 61.9\pmstd{0.3}& 13{,}839 & 264.8 \\
VGSC $\alpha=10$ & \underline{81.1}\pmstd{0.4}& \underline{45.0}\pmstd{0.3}& \textbf{70.3}\pmstd{0.3}& \textbf{54.5}\pmstd{0.4}& \textbf{64.5}\pmstd{1.1}& \textbf{63.1}\pmstd{0.3}& 15{,}179 & 290.2 \\
VGSC $\alpha=20$ & 80.6\pmstd{0.5}& \textbf{45.4}\pmstd{0.3}& 69.6\pmstd{0.1}& \underline{54.4}\pmstd{1.1}& 62.5\pmstd{1.2}& \underline{62.5}\pmstd{0.3}& 16{,}093 & 307.6 \\
    \bottomrule
  \end{tabular}
  }
\label{tab:qwen_prm}
  
\end{table*}

  Table~\ref{tab:qwen_prm} reports the main results for \qwen with \qwenprm verifiers; Appendix~\ref{app:additional_results} reports results for other LLMs (\llama and \gemma) and verifier families (\reprobe).

  Overall, VGSC achieves the strongest average performance among the evaluated decoding strategies. VGSC with $\alpha=10$ achieves the highest average accuracy, $63.1\%$, exceeding the strongest non-VGSC baseline, BS+SC-Future, by $2.4$ percentage points and WSC with a similar nominal budget of $N=25$ by $3.7$ points. Across its variants, VGSC achieves or ties the best result on all five benchmarks. The three settings $\alpha\in\{5,10,20\}$ achieve the three highest average accuracies, suggesting that moderate-to-strong verifier guidance is effective in this setting.
 
  Table~\ref{tab:qwen_prm} also reports generated tokens and estimated inference compute per MATH example. VGSC can accept a  proposed step before exhausting its candidate budget, reducing the number of generated tokens. For example, VGSC with $\alpha=5$ achieves $81.2\%$ accuracy using $13{,}839$ tokens per example, compared with $77.4\%$ accuracy and $19{,}837$ tokens for BS-Step, a $30.2\%$ token reduction. This saving is baseline-dependent: BS-Future generates fewer tokens but achieves lower accuracy.

  The reported TFLOPs follow Eq.~\eqref{eq:compute_formula} under a conditional shared-prefix implementation. Because the two resulting PRMs differ only in their final four layers, this estimate counts a shared frozen prefix and two four-layer branches, corresponding to $8$B effective verifier parameters under an equal-layer approximation. Section~\ref{sec:ablation_budget} further examines the accuracy--compute tradeoff across search budgets, and Appendix~\ref{app:inference_compute} details the compute estimates.

\subsection{Ablations}
\label{sec:ablations}

  We examine how VGSC scales with search budget, how it responds to corrupted verifier scores, and how its performance depends on the step-level verifier $V$ and future-outcome predictor $\widehat{Q}$. Additional appendix ablations examine score aggregation (Appendix~\ref{app:alternative_aggregations}), the verifier-trust exponent (Appendix~\ref{app:ablation_alpha}), and allocation under a fixed nominal budget (Appendix~\ref{app:ablation_fixed_nc}). Appendix~\ref{app:analysis} provides case studies, coverage statistics, and diversity statistics.

  \begin{figure*}[t]
    \centering
    \includegraphics[width=\textwidth, trim=0cm 0.2cm 0cm 0cm, clip]{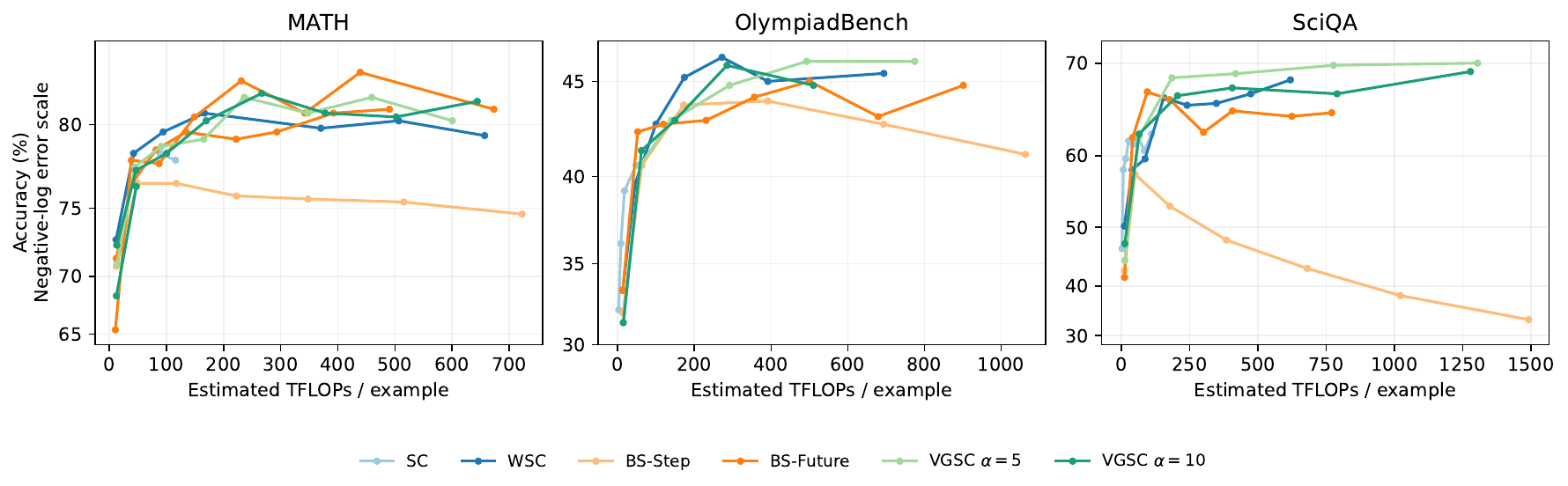}
    \caption{
      Accuracy--compute tradeoff on MATH, OlympiadBench, and SciQA with Qwen2.5-Math-1.5B and PRM-7B verifiers. For search methods, the budget increases as $N=C=k$, jointly scaling cross-trajectory exploration and within-step candidate generation; SC and WSC use $k^2$ trajectories. The horizontal axis reports inference compute in TFLOPs per example.
    }
    \label{fig:ablation_budget_equal_nc}
    
  \end{figure*}

  \begin{figure*}[!t]
    \centering
    \includegraphics[width=\textwidth, trim=0cm 0.2cm 0cm 0cm, clip]{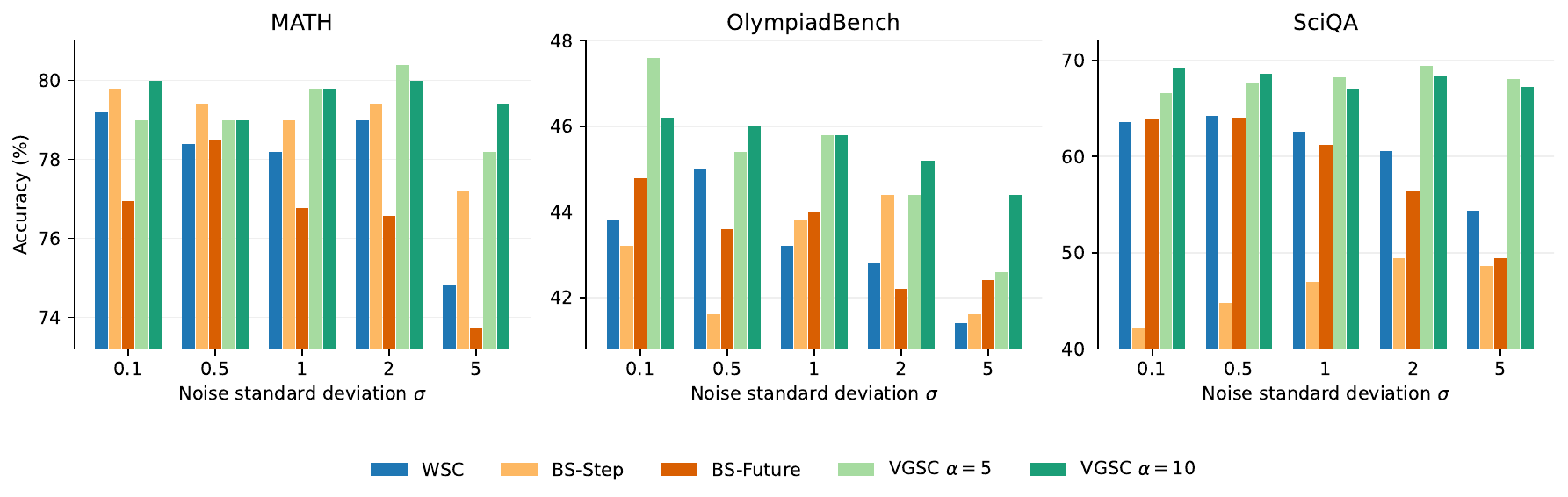}
    \caption{Robustness to corrupted verifier scores. We perturb verifier outputs by adding Gaussian noise in logit space with standard deviation $\sigma$ and then mapping back to probabilities.}
    \label{fig:ablation_verifier_noise}
    
  \end{figure*}

\paragraph{Scaling total budget with balanced search.}
\label{sec:ablation_budget}
  We study budget scaling on random subsets of 100 examples from MATH, OlympiadBench, and SciQA. Figure~\ref{fig:ablation_budget_equal_nc} studies the compute-accuracy trade-off in a balanced regime $N=C$ ranging $N$ from 1 to 8, where increasing $N$ raises both the number of independent trajectories and verifier-guided candidate expansions per step. Beam-search baselines remain competitive on MATH, likely because the PRM verifier acts in-domain and is less prone to mis-scoring. On the verifier out-of-domain datasets, OlympiadBench and SciQA, BS variants degrade as the budget grows, starting around $N \geq 5$ on OlympiadBench and $N \geq 3$ on SciQA for BS-Step. This matches~\citet{dalal2026testtimecomputehurtoverestimation}: increasing test-time compute can hurt when search aggressively maximizes noisy verifier scores. In contrast, sampling-based methods are more stable, and VGSC remains among the top-performing methods across nearly all budgets.

\paragraph{Robustness to corrupted verifier scores.}
  We evaluate robustness on the same random subsets of 100 examples from MATH, OlympiadBench, and SciQA. We perturb verifier logits with Gaussian noise of standard deviation $\sigma$ before mapping them back to probabilities with a sigmoid, corrupting both $V$ and $\widehat{Q}$ for methods that use both signals. Figure~\ref{fig:ablation_verifier_noise} shows that accuracy generally decreases as noise increases, although the effect varies across methods and datasets. VGSC remains comparatively stable, particularly on SciQA, where beam-search baselines degrade more substantially. These results suggest that joint verifier guidance combined with stochastic exploration can reduce sensitivity to corrupted scores.

\paragraph{Dependence on verifier components.}
  We examine how VGSC's performance depends on the step-level verifier $V$ and future-outcome predictor $\widehat{Q}$ on the five main evaluation benchmarks. For VGSC, we compare joint guidance $(V\widehat{Q})^\alpha$ with $V^\alpha$ alone and $\widehat{Q}^\alpha$ alone, keeping the remaining decoding settings unchanged. We also test whether combining the signals benefits deterministic search. For BS and BS+SC, we replace the step-only score with
  \begin{equation}
  W(y_t \mid \xv,\yv_{<t})=V(y_t \mid \xv,\yv_{<t})\widehat{Q}(\yv_{\le t},\xv).
  \end{equation}

\begin{wraptable}{r}{0.35\textwidth}
  \centering
  \small
  \setlength{\tabcolsep}{4pt}
  \vspace{-1em}
  \caption{Effect of verifier components for \qwen with \qwenprm. Accuracy is averaged across five benchmarks. Bold entries identify the best VGSC variant within each $\alpha$ group.}
  \begin{tabular}{llcc}
    \toprule
    Method & Guidance & $\alpha$ & Avg. acc.\ $\uparrow$ \\
    \midrule
    BS & $V$ & --- & 46.0 \\
    BS & $V\widehat{Q}$ & --- & 53.6 \\
    BS+SC & $V$ & --- & 57.0 \\
    BS+SC & $V\widehat{Q}$ & --- & 60.8 \\
    \midrule
    VGSC & $V$ only & 5 & 61.0 \\
    VGSC & $\widehat{Q}$ only & 5 & \underline{57.5} \\
    VGSC & $V\widehat{Q}$ & 5 & \textbf{61.9} \\
    \midrule
    VGSC & $V$ only & 10 & 61.8 \\
    VGSC & $\widehat{Q}$ only & 10 & \underline{62.4} \\
    VGSC & $V\widehat{Q}$ & 10 & \textbf{63.1} \\
    \bottomrule
  \end{tabular}
  \label{tab:ablation_alternative_baselines}
  \vspace{-2em}
\end{wraptable}
  A positive exponent applied to the entire joint score preserves its ranking and therefore does not affect beam selection.

  Table~\ref{tab:ablation_alternative_baselines} shows that joint guidance improves average accuracy for BS from $46.0\%$ to $53.6\%$ and for BS+SC from $57.0\%$ to $60.8\%$. Both remain below full VGSC, which achieves $61.9\%$ at $\alpha=5$ and $63.1\%$ at $\alpha=10$. Within VGSC, removing either component reduces average accuracy at both trust settings. These results suggest that performance depends both on the available verifier signals and on how they guide decoding.

\section{Related Work}


\paragraph{Verification for reasoning.}
  Prior work has improved multi-step reasoning by training verifiers to assess either complete solutions or intermediate reasoning steps. These include outcome-level verifiers and process reward models (PRMs) for step-level supervision~\citep{cobbe2021gsm8k, lightman2023letsverifystepstep, wang-etal-2024-math}. Recent work further studies how verifier quality depends on supervision, evaluation design, and verifier architecture~\citep{prmlessons, ni2026efficienttesttimescalingmultistep}. Our focus is complementary: rather than proposing a new verifier, we study how verifier signals should be used at inference time.

\paragraph{Continuation values and progress-based verification.}
  Step correctness and continuation value provide different forms of guidance: a locally valid step need not make the remaining problem easier, while the probability of successful completion depends on both the prefix and the policy used to continue it. \citet{setlur2024rewardingprogressscalingautomated} study process-advantage verifiers that assess progress toward successful completion, highlighting the importance of policy-dependent continuation estimates. VGSC similarly uses future-outcome information, but retains a separate step-correctness verifier. Our contribution concerns how these two signals enter a stochastic decoding rule, rather than a new value-estimation objective.

\paragraph{Inference-time verification.}
  A common use of test-time compute is post-hoc aggregation: multiple reasoning trajectories are sampled from the base model and then combined by majority vote or verifier-weighted reranking~\citep{wang2023selfconsistencyimproveschainthought, cobbe2021gsm8k, kang2025scalablebestofnselectionlarge}. These methods preserve diversity, but apply verification only after generation is complete. Other methods inject verifier information during decoding, including tree/search-based reasoning~\citep{yao2023treethoughtsdeliberateproblem}, step-level discriminator guidance~\citep{khalifa-etal-2023-grace}, and self-evaluation guided stochastic beam search~\citep{xie2023selfevaluationguidedbeamsearch}. These approaches use local feedback to guide generation, but do not explicitly model the future success probability of continuing from a partial trajectory. Future-looking methods guide search using estimates of continuation quality. Value-guided search (VGS) uses a learned continuation-value model to guide block-wise search and aggregates completed trajectories through value-weighted majority voting~\citep{Wang2025ValueGuidedSF}. This approach is closely related to our $\widehat{Q}$-only variant, which isolates future-outcome guidance within our decoding framework. VGSC additionally incorporates step-level correctness and uses stochastic acceptance/resampling rather than highest-value candidate selection. Twisted sequential Monte Carlo (TSMC) samples from a base language model tilted by a sequence-level potential using learned twists, particle reweighting, and resampling~\citep{zhao2024probabilisticinferencelanguagemodels}. When this potential is a verifier score, TSMC targets a verifier-tilted distribution $P^V_1$, but it treats the verifier as a terminal sequence-level signal and relies on learned twists for intermediate guidance. VGSC instead exploits intermediate process-verifier scores directly, combining $V$ and $\widehat{Q}$ in each local next-step resampling decision while preserving multiple trajectories for final self-consistency.

\paragraph{Imperfect verifier signals.}
  Learned verifiers are imperfect: PRMs can be miscalibrated, overestimate partial trajectory quality, and be exploited when optimized too aggressively at inference time~\citep{park2025knowdontknowuncertainty, khalaf2025inferencetimerewardhackinglarge}. More broadly, search methods that rely heavily on verifier scores can suffer from over-optimistic selection, and wider verifier-guided search need not monotonically improve performance~\citep{dalal2026testtimecomputehurtoverestimation, cinquin2025limitsprmguidedtreesearch}. This motivates our emphasis on stochastic guidance that balances generator exploration with verifier information, rather than deterministic maximization of noisy scores.


\section{Conclusion}
  We introduced Verifier-Guided Self-Consistency (VGSC), a test-time reasoning method that combines generator sampling with verifier guidance during decoding. VGSC uses both step-level correctness estimates and future-outcome prediction to guide local sampling, while still producing multiple diverse reasoning trajectories for final self-consistency. This yields a simple alternative to both post-hoc verifier reranking and beam-style maximization.

  Across multiple reasoning benchmarks, generator models, and verifier families, our results show that VGSC consistently improves over representative post-hoc aggregation baselines and achieves a stronger balance between exploration and verifier guidance than beam-style search. These findings suggest that inference-time reasoning benefits from stochastic guidance toward promising trajectories rather than aggressive deterministic pruning.

\section*{Limitations}
  VGSC requires auxiliary learned guidance signals at inference time, in particular the future-outcome predictor $\widehat{Q}$. In our implementation, $\widehat{Q}$ is trained from Monte Carlo continuation outcomes for a fixed generator/verifier configuration, which adds some training and engineering overhead relative to simpler decoding schemes that only reuse an existing verifier at test time. We view this as a moderate limitation rather than an unusual requirement: Monte Carlo continuation correctness signals are already commonly used to train PRMs, since they can be obtained automatically without human step-level annotations. VGSC repurposes this same kind of automatically generated supervision to estimate continuation value, and in our experiments this additional signal provides a useful complement to step-level verification.

  More broadly, VGSC depends on the quality and task alignment of the learned guidance signals. As with other verifier-guided methods, gains may be smaller when the verifier is weak, miscalibrated, or poorly matched to the target domain. Our results on CommonsenseQA with \reprobe illustrate this regime, where verifier-based guidance is less helpful than in mathematical and logical settings.

  Finally, VGSC introduces the verifier-trust exponent $\alpha$ as a method-specific hyperparameter. Although we find that a default range of $\alpha \in [10,20]$ works well across datasets, the best setting may vary with verifier quality, generator behavior, and task domain. This suggests that some validation or tuning may be needed when applying VGSC to new settings.
\section*{Ethical Considerations}
  VGSC aims to improve the reliability of language-model reasoning by using verifier guidance during test-time generation. However, it does not guarantee correctness: generated reasoning can still be flawed, and final answers can still be wrong.

  The method depends on learned verifiers and future-outcome predictors, whose scores may be biased, miscalibrated, or poorly aligned with new domains. If these models assign high confidence to incorrect steps, VGSC may amplify such errors, especially under strong verifier guidance. Careful calibration, monitoring, and domain-specific validation are therefore important for practical deployment.

\clearpage

\bibliographystyle{tmlr}
\bibliography{references}

@article{kang2025scalablebestofnselectionlarge,
  title={Scalable best-of-n selection for large language models via self-certainty},
  author={Kang, Zhewei and Zhao, Xuandong and Song, Dawn},
  journal={Advances in neural information processing systems},
  volume={38},
  pages={19720--19745},
  year={2026}
}

@inproceedings{shelmanov-etal-2025-head,
  title={A head to predict and a head to question: Pre-trained uncertainty quantification heads for hallucination detection in llm outputs},
  author={Shelmanov, Artem and Fadeeva, Ekaterina and Tsvigun, Akim and Tsvigun, Ivan and Xie, Zhuohan and Kiselev, Igor and Daheim, Nico and Zhang, Caiqi and Vazhentsev, Artem and Sachan, Mrinmaya and others},
  booktitle={Proceedings of the 2025 Conference on Empirical Methods in Natural Language Processing},
  pages={35700--35719},
  year={2025}
}

@article{wang2023selfconsistencyimproveschainthought,
  title={Self-consistency improves chain of thought reasoning in language models},
  author={Wang, Xuezhi and Wei, Jason and Schuurmans, Dale and Le, Quoc and Chi, Ed and Narang, Sharan and Chowdhery, Aakanksha and Zhou, Denny},
  journal={arXiv preprint arXiv:2203.11171},
  year={2022}
}

@article{Wang2025ValueGuidedSF,
  title={Value-Guided Search for Efficient Chain-of-Thought Reasoning},
  author={Kaiwen Wang and Jin-Peng Zhou and Jonathan D. Chang and Zhao-Lin Gao and Nathan Kallus and Kiant{\'e} Brantley and Wen Sun},
  journal={ArXiv},
  year={2025},
  volume={abs/2505.17373},
  url={https://api.semanticscholar.org/CorpusID:278886339}
}

@article{sympy,
  title={SymPy: symbolic computing in Python},
  author={Meurer, Aaron and Smith, Christopher P and Paprocki, Mateusz and {\v{C}}ert{\'\i}k, Ond{\v{r}}ej and Kirpichev, Sergey B and Rocklin, Matthew and Kumar, AMiT and Ivanov, Sergiu and Moore, Jason K and Singh, Sartaj and others},
  journal={PeerJ Computer Science},
  volume={3},
  pages={e103},
  year={2017},
  publisher={PeerJ Inc.}
}

@article{xie2023selfevaluationguidedbeamsearch,
  title={Self-Evaluation Guided Beam Search for Reasoning},
  author={Yuxi Xie and Kenji Kawaguchi and Yiran Zhao and Xu Zhao and MingSung Kan and Junxian He and Qizhe Xie},
  journal={Advances in Neural Information Processing Systems 36},
  year={2023}
}

@article{grattafiori2024llama,
  title={The llama 3 herd of models},
  author={Grattafiori, Aaron and Dubey, Abhimanyu and Jauhri, Abhinav and Pandey, Abhinav and Kadian, Abhishek and Al-Dahle, Ahmad and Letman, Aiesha and Mathur, Akhil and Schelten, Alan and Vaughan, Alex and others},
  journal={arXiv preprint arXiv:2407.21783},
  year={2024}
}

@article{zhao2024probabilisticinferencelanguagemodels,
  title={Probabilistic inference in language models via twisted sequential monte carlo},
  author={Zhao, Stephen and Brekelmans, Rob and Makhzani, Alireza and Grosse, Roger},
  journal={arXiv preprint arXiv:2404.17546},
  year={2024}
}

@article{Kaplan2020ScalingLF,
  title={Scaling laws for neural language models},
  author={Kaplan, Jared and McCandlish, Sam and Henighan, Tom and Brown, Tom B and Chess, Benjamin and Child, Rewon and Gray, Scott and Radford, Alec and Wu, Jeffrey and Amodei, Dario},
  journal={arXiv preprint arXiv:2001.08361},
  year={2020}
}

@article{Agapiou2015ImportanceSI,
  title={Importance sampling: Intrinsic dimension and computational cost},
  author={Agapiou, Sergios and Papaspiliopoulos, Omiros and Sanz-Alonso, Daniel and Stuart, Andrew M},
  journal={Statistical Science},
  pages={405--431},
  year={2017},
  publisher={JSTOR}
}

@inproceedings{lin2004rouge,
  title={Rouge: A package for automatic evaluation of summaries},
  author={Lin, Chin-Yew},
  booktitle={Text summarization branches out},
  pages={74--81},
  year={2004}
}

@article{yao2023treethoughtsdeliberateproblem,
  title={Tree of thoughts: Deliberate problem solving with large language models},
  author={Yao, Shunyu and Yu, Dian and Zhao, Jeffrey and Shafran, Izhak and Griffiths, Tom and Cao, Yuan and Narasimhan, Karthik},
  journal={Advances in neural information processing systems},
  volume={36},
  pages={11809--11822},
  year={2023}
}

@article{ni2026efficienttesttimescalingmultistep,
  title={Efficient Test-Time Scaling of Multi-Step Reasoning by Probing Internal States of Large Language Models},
  author={Ni, Jingwei and Fadeeva, Ekaterina and Wu, Tianyi and Akhtar, Mubashara and Zhang, Jiaheng and Ash, Elliott and Leippold, Markus and Baldwin, Timothy and Ng, See-Kiong and Shelmanov, Artem and others},
  journal={arXiv preprint arXiv:2511.06209},
  year={2025}
}

@inproceedings{khalifa-etal-2023-grace,
  title={Grace: Discriminator-guided chain-of-thought reasoning},
  author={Khalifa, Muhammad and Logeswaran, Lajanugen and Lee, Moontae and Lee, Honglak and Wang, Lu},
  booktitle={Findings of the Association for Computational Linguistics: EMNLP 2023},
  pages={15299--15328},
  year={2023}
}

@inproceedings{wang-etal-2024-math,
  title={Math-shepherd: Verify and reinforce llms step-by-step without human annotations},
  author={Wang, Peiyi and Li, Lei and Shao, Zhihong and Xu, Runxin and Dai, Damai and Li, Yifei and Chen, Deli and Wu, Yu and Sui, Zhifang},
  booktitle={Proceedings of the 62nd Annual Meeting of the Association for Computational Linguistics (Volume 1: Long Papers)},
  pages={9426--9439},
  year={2024}
}

@inproceedings{lightman2023letsverifystepstep,
  title={Let's verify step by step},
  author={Lightman, Hunter and Kosaraju, Vineet and Burda, Yuri and Edwards, Harrison and Baker, Bowen and Lee, Teddy and Leike, Jan and Schulman, John and Sutskever, Ilya and Cobbe, Karl},
  booktitle={International Conference on Learning Representations},
  volume={2024},
  pages={39578--39601},
  year={2024}
}

@article{cobbe2021gsm8k,
  title={Training verifiers to solve math word problems, 2021},
  author={Cobbe, Karl and Kosaraju, Vineet and Bavarian, Mohammad and Chen, Mark and Jun, Heewoo and Kaiser, Lukasz and Plappert, Matthias and Tworek, Jerry and Hilton, Jacob and Nakano, Reiichiro and others},
  journal={URL https://arxiv. org/abs/2110.14168},
  volume={9},
  year={2021}
}

@article{wei2023chainofthoughtpromptingelicitsreasoning,
  title={Chain-of-thought prompting elicits reasoning in large language models},
  author={Wei, Jason and Wang, Xuezhi and Schuurmans, Dale and Bosma, Maarten and Xia, Fei and Chi, Ed and Le, Quoc V and Zhou, Denny and others},
  journal={Advances in neural information processing systems},
  volume={35},
  pages={24824--24837},
  year={2022}
}

@article{hu2021loralowrankadaptationlarge,
  title={Lora: Low-rank adaptation of large language models.},
  author={Hu, Edward J and Shen, Yelong and Wallis, Phillip and Allen-Zhu, Zeyuan and Li, Yuanzhi and Wang, Shean and Wang, Liang and Chen, Weizhu and others},
  journal={Iclr},
  volume={1},
  number={2},
  pages={3},
  year={2022}
}

@inproceedings{talmor-etal-2019-commonsenseqa,
  title={Commonsenseqa: A question answering challenge targeting commonsense knowledge},
  author={Talmor, Alon and Herzig, Jonathan and Lourie, Nicholas and Berant, Jonathan},
  booktitle={Proceedings of the 2019 Conference of the North American Chapter of the Association for Computational Linguistics: Human Language Technologies, Volume 1 (Long and Short Papers)},
  pages={4149--4158},
  year={2019}
}

@inproceedings{prmlessons,
  title={The lessons of developing process reward models in mathematical reasoning},
  author={Zhang, Zhenru and Zheng, Chujie and Wu, Yangzhen and Zhang, Beichen and Lin, Runji and Yu, Bowen and Liu, Dayiheng and Zhou, Jingren and Lin, Junyang},
  booktitle={Findings of the Association for Computational Linguistics: ACL 2025},
  pages={10495--10516},
  year={2025}
}

@inproceedings{xie2024memorization,
  title={On memorization of large language models in logical reasoning},
  author={Xie, Chulin and Huang, Yangsibo and Zhang, Chiyuan and Yu, Da and Chen, Xinyun and Lin, Bill Yuchen and Li, Bo and Ghazi, Badih and Kumar, Ravi},
  booktitle={Proceedings of the 14th International Joint Conference on Natural Language Processing and the 4th Conference of the Asia-Pacific Chapter of the Association for Computational Linguistics},
  pages={2742--2785},
  year={2025}
}

@article{cinquin2025limitsprmguidedtreesearch,
  title={Limits of PRM-Guided Tree Search for Mathematical Reasoning with LLMs},
  author={Cinquin, Tristan and Pleiss, Geoff and Kristiadi, Agustinus},
  journal={arXiv preprint arXiv:2510.20272},
  year={2025}
}

@article{dalal2026testtimecomputehurtoverestimation,
  title={More Test-Time Compute Can Hurt: Overestimation Bias in LLM Beam Search},
  author={Dalal, Gal and Hallak, Assaf and Chechik, Gal and Ziser, Yftah},
  journal={arXiv preprint arXiv:2603.15377},
  year={2026}
}

@article{khalaf2025inferencetimerewardhackinglarge,
  title={Inference-time reward hacking in large language models},
  author={Khalaf, Hadi and Mayrink Verdun, Claudio and Oesterling, Alex and Lakkaraju, Himabindu and Calmon, Flavio},
  journal={Advances in Neural Information Processing Systems},
  volume={38},
  pages={61720--61760},
  year={2026}
}

@inproceedings{setlur2024rewardingprogressscalingautomated,
  title={Rewarding progress: Scaling automated process verifiers for llm reasoning},
  author={Setlur, Amrith and Nagpal, Chirag and Fisch, Adam and Geng, Xinyang and Eisenstein, Jacob and Agarwal, Rishabh and Agarwal, Alekh and Berant, Jonathan and Kumar, Aviral},
  booktitle={International Conference on Learning Representations},
  volume={2025},
  pages={60808--60838},
  year={2025}
}

@article{park2025knowdontknowuncertainty,
  title={Know What You Don't Know: Uncertainty Calibration of Process Reward Models},
  author={Park, Young-Jin and Greenewald, Kristjan and Alimohammadi, Kaveh and Wang, Hao and Azizan, Navid},
  journal={Advances in Neural Information Processing Systems},
  volume={38},
  pages={38852--38895},
  year={2026}
}

@inproceedings{yu-etal-2024-ovm,
  title={Ovm, outcome-supervised value models for planning in mathematical reasoning},
  author={Yu, Fei and Gao, Anningzhe and Wang, Benyou},
  booktitle={Findings of the Association for Computational Linguistics: NAACL 2024},
  pages={858--875},
  year={2024}
}

@inproceedings{hendrycksmath2021,
  title={Measuring Mathematical Problem Solving With the MATH Dataset},
  author={Hendrycks, Dan and Burns, Collin and Kadavath, Saurav and Arora, Akul and Basart, Steven and Tang, Eric and Song, Dawn and Steinhardt, Jacob},
  booktitle={Thirty-fifth Conference on Neural Information Processing Systems Datasets and Benchmarks Track (Round 2)},
  year={2021}
}

@inproceedings{He2024OlympiadBenchAC,
  title={Olympiadbench: A challenging benchmark for promoting agi with olympiad-level bilingual multimodal scientific problems},
  author={He, Chaoqun and Luo, Renjie and Bai, Yuzhuo and Hu, Shengding and Thai, Zhen and Shen, Junhao and Hu, Jinyi and Han, Xu and Huang, Yujie and Zhang, Yuxiang and others},
  booktitle={Proceedings of the 62nd Annual Meeting of the Association for Computational Linguistics (Volume 1: Long Papers)},
  pages={3828--3850},
  year={2024}
}

@article{lu2022learn,
  title={Learn to explain: Multimodal reasoning via thought chains for science question answering},
  author={Lu, Pan and Mishra, Swaroop and Xia, Tanglin and Qiu, Liang and Chang, Kai-Wei and Zhu, Song-Chun and Tafjord, Oyvind and Clark, Peter and Kalyan, Ashwin},
  journal={Advances in neural information processing systems},
  volume={35},
  pages={2507--2521},
  year={2022}
}

@article{yang2024qwen25mathtechnicalreportmathematical,
  title={Qwen2. 5-math technical report: Toward mathematical expert model via self-improvement},
  author={Yang, An and Zhang, Beichen and Hui, Binyuan and Gao, Bofei and Yu, Bowen and Li, Chengpeng and Liu, Dayiheng and Tu, Jianhong and Zhou, Jingren and Lin, Junyang and others},
  journal={arXiv preprint arXiv:2409.12122},
  year={2024}
}

@article{Kamath2025Gemma3T,
  title={Gemma 3 Technical Report},
  author={Gemma Team Aishwarya Kamath and Johan Ferret and Shreya Pathak and Nino Vieillard and Ramona Merhej and Sarah Perrin and Tatiana Matejovicova and Alexandre Ram'e and Morgane Rivi{\`e}re and Louis Rouillard and Thomas Mesnard and Geoffrey Cideron and Jean-Bastien Grill and Sabela Ramos and Edouard Yvinec and Michelle Casbon and Etienne Pot and Ivo Penchev and Gael Liu and Francesco Visin and Kathleen Kenealy and Lucas Beyer and Xiaohai Zhai and Anton Tsitsulin and R{\'o}bert Istvan Busa-Fekete and Alex Feng and Noveen Sachdeva and Benjamin Coleman and Yi Gao and Basil Mustafa and Iain Barr and Emilio Parisotto and David Tian and Matan Eyal and Colin Cherry and Jan-Thorsten Peter and Danila Sinopalnikov and Surya Bhupatiraju and Rishabh Agarwal and Mehran Kazemi and Dan Malkin and Ravin Kumar and David Vilar and Idan Brusilovsky and Jiaming Luo and Andreas Steiner and Abe Friesen and Abhanshu Sharma and Abheesht Sharma and Adi Mayrav Gilady and Adrian Goedeckemeyer and Alaa Saade and Alexander Kolesnikov and Alexei Bendebury and Alvin Abdagic and Amit Vadi and Andr'as Gyorgy and Andr{\'e} Susano Pinto and Anil Das and Ankur Bapna and Antoine Miech and Antoine Yang and Antonia Paterson and Ashish Shenoy and Ayan Chakrabarti and Bilal Piot and Boxi Wu and Bobak Shahriari and Bryce Petrini and Charlie Chen and Charline Le Lan and Christopher A. Choquette-Choo and Cj Carey and Cormac Brick and Daniel Deutsch and Danielle Eisenbud and Dee Cattle and Derek Zhiyuan Cheng and Dimitris Paparas and Divyashree Shivakumar Sreepathihalli and Doug Reid and Dustin Tran and Dustin Zelle and Eric Noland and Erwin Huizenga and Eugene Kharitonov and Frederick Liu and Gagik Amirkhanyan and Glenn Cameron and Hadi Hashemi and Hanna Klimczak-Pluci'nska and Harman Singh and Harsh Mehta and Harshal Tushar Lehri and Hussein Hazimeh and Ian Ballantyne and Idan Szpektor and Ivan Nardini and Jean Pouget-Abadie and Jetha Chan and Joe Stanton and J. Michael Wieting and Jonathan Lai and Jordi Orbay and Joe Fernandez and Joshua Newlan and Junsong Ji and Jyotinder Singh and Kat Black and Kathy Yu and Kevin Hui and Kiran Vodrahalli and Klaus Greff and Linhai Qiu and Marcella Valentine and Marina Coelho and Marvin Ritter and Matt Hoffman and Matthew Watson and Mayank Chaturvedi and Michael Moynihan and Min Ma and Nabila Babar and Natasha Noy and Nathan Byrd and Nick Roy and Nikola Momchev and Nilay Chauhan and Oskar Bunyan and Pankil Botarda and Paul Caron and Paul Kishan Rubenstein and Phil Culliton and Philipp Schmid and Pier Giuseppe Sessa and Ping-mei Xu and Piotr Stańczyk and Pouya Dehghani Tafti and Rakesh Shivanna and Renjie Wu and Renke Pan and Reza Ardeshir Rokni and Rob Willoughby and Rohith Vallu and Ryan Mullins and Sammy Jerome and Sara Smoot and Sertan Girgin and Shariq Iqbal and Shashir Reddy and Shruti Sheth and Siim P{\~o}der and Sijal Bhatnagar and Sindhu Raghuram Panyam and Sivan Eiger and Susan Zhang and Tianqi Liu and Trevor Yacovone and Tyler Liechty and Uday Kalra and Utku Evci and Vedant Misra and Vincent Roseberry and Vladimir Feinberg and Vlad Kolesnikov and Woohyun Han and Woosuk Kwon and Xi Chen and Yinlam Chow and Yuvein Zhu and Zichuan Wei and Zoltan Egyed and Victor Cotruta and Minh Giang and Phoebe Kirk and Anand Rao and Jessica Lo and Erica Moreira and Luiz Gustavo Martins and Omar Sanseviero and Lucas Gonzalez and Zach Gleicher and Tris Warkentin and Vahab S. Mirrokni and Evan Senter and Eli Collins and Joelle Barral and Zoubin Ghahramani and Raia Hadsell and Yossi Matias and D. Sculley and Slav Petrov and Noah Fiedel and Noam Shazeer and Oriol Vinyals and Jeffrey Dean and Demis Hassabis and Koray Kavukcuoglu and Cl{\'e}ment Farabet and Elena Buchatskaya and Jean-Baptiste Alayrac and Rohan Anil and Dmitry Lepikhin and Sebastian Borgeaud and Olivier Bachem and Armand Joulin and Alek Andreev and Cassidy Hardin and Robert Dadashi and L'eonard Hussenot},
  journal={ArXiv},
  year={2025},
  volume={abs/2503.19786}
}

@misc{thinkbooster2026,
  title     = {ThinkBooster: A Unified Framework for Seamless Test-Time Scaling of LLM Reasoning},
  author    = {Smirnov, Vladislav and Nguyen, Chieu and Senichev, Sergey and Ta, Minh Ngoc and Fadeeva, Ekaterina and Vazhentsev, Artem and Galimzianova, Daria and Rozanov, Nikolai and Mazanov, Viktor and Ni, Jingwei and Wu, Tianyi and Kiselev, Igor and Sachan, Mrinmaya and Gurevych, Iryna and Nakov, Preslav and Baldwin, Timothy and Shelmanov, Artem},
  booktitle = {Preprint},
  year      = {2026}
}

@article{li_efficient_2026,
  title={Efficient Test-Time Inference via Deterministic Exploration of Truncated Decoding Trees},
  author={Li, Xueyan and Zenn, Johannes and Fadeeva, Ekaterina and Su, Guinan and Sachan, Mrinmaya and Geiping, Jonas},
  journal={arXiv preprint arXiv:2604.20500},
  year={2026}
}

\clearpage

\newpage
\appendix

\section{Additional Ablation Studies}
\label{app:ablations}
  This section presents supplementary ablation experiments that examine the robustness of our results to key design choices. In particular, we study alternative ways of aggregating step-level verifier signals and evaluate additional baseline variants that make use of verifier information differently.

\subsection{Alternative Aggregations}
\label{app:alternative_aggregations}
  
  We study whether verifier-guided methods depend on how step-level verifier scores are aggregated along a reasoning chain. All default method variants in the main paper use \textit{product} aggregation; here, we additionally evaluate \textit{min} and \textit{mean}. Specifically, for the \qwen + \qwenprm setting, we consider WSC, DLE, BS-Step, and BS+SC-Step variants in which the default product score is replaced by one of these alternatives. For a partial trajectory $\yv_{\le t}$, we define the aggregated verifier score as
  \begin{align}
    V^\text{agg}(\yv_{\le t} \mid \xv) = \operatorname{agg}(V&(y_1 \mid \xv), \dots, V(y_t \mid \xv, \yv_{<t})),
  \end{align}
  where $\operatorname{agg} \in \{\text{product}, \text{min}, \text{mean}\}$. Product favors consistently high step scores, min is determined by the weakest step, and mean provides a softer notion of average prefix quality.
  
  For VGSC, we analogously aggregate both the step-level verifier scores and future-outcome score into a single guidance signal,
  \begin{align}
      W^\text{agg}_\alpha(y_t \mid \xv, \yv_{<t}) = \operatorname{agg}\!\left(V(y_1 \mid \xv), .., V(y_t \mid \xv, \yv_{<t}), \widehat{Q}(\yv_{\le t}, \xv)\right)^\alpha,
  \end{align}
  where product is the default VGSC variant and min/mean are the alternatives studied here. We evaluate these variants with $\alpha = 10$, the best setting from the main experiments.
  
  Table~\ref{tab:ablation_aggregations} shows that aggregation choice has only a small effect on average accuracy. Product remains a strong default, while the comparable performance of min and mean suggests that verifier-guided decoding is not highly sensitive to this choice.



\subsection{Verifier-Trust Parameter Sensitivity}
\label{app:ablation_alpha}

\begin{figure*}[h]
    \centering
    \includegraphics[width=0.95\textwidth, trim=0cm 0.3cm 0cm 0cm, clip]{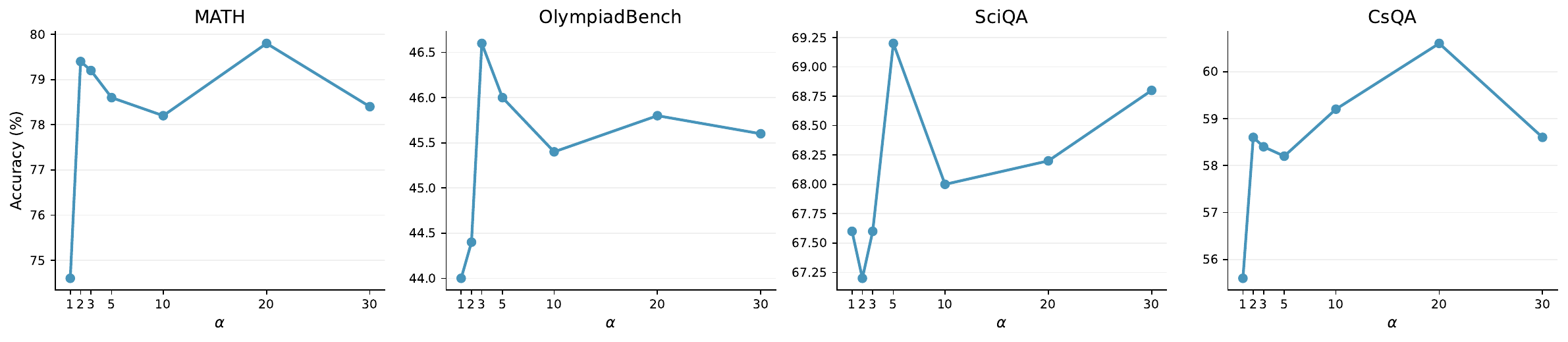}
    \vspace{-1em}
    \caption{VGSC sensitivity to the verifier-trust exponent $\alpha$ for $100$ sample subsets of 3 datasets.}
    \label{fig:ablation_alpha}
  \end{figure*}

Figure~\ref{fig:ablation_alpha} examines the verifier-trust exponent $\alpha$, which controls the strength of verifier guidance during decoding. Across the evaluated datasets, increasing $\alpha$ from small values generally improves accuracy, with the strongest performance typically attained around $\alpha=3$--$5$. Further increases provide limited additional benefit and can reduce accuracy, suggesting that stronger guidance may amplify errors in the verifier signals. These results favor moderate verifier trust, although the best setting varies across datasets. We therefore treat $\alpha$ as a task-dependent hyperparameter rather than recommend a single range for all tasks.

\subsection{Compute Allocation under Fixed Budget}
\label{app:ablation_fixed_nc}
\begin{figure*}[!h]
    \centering
    \includegraphics[width=0.95\textwidth, trim=0cm 0.3cm 0cm 0cm, clip]{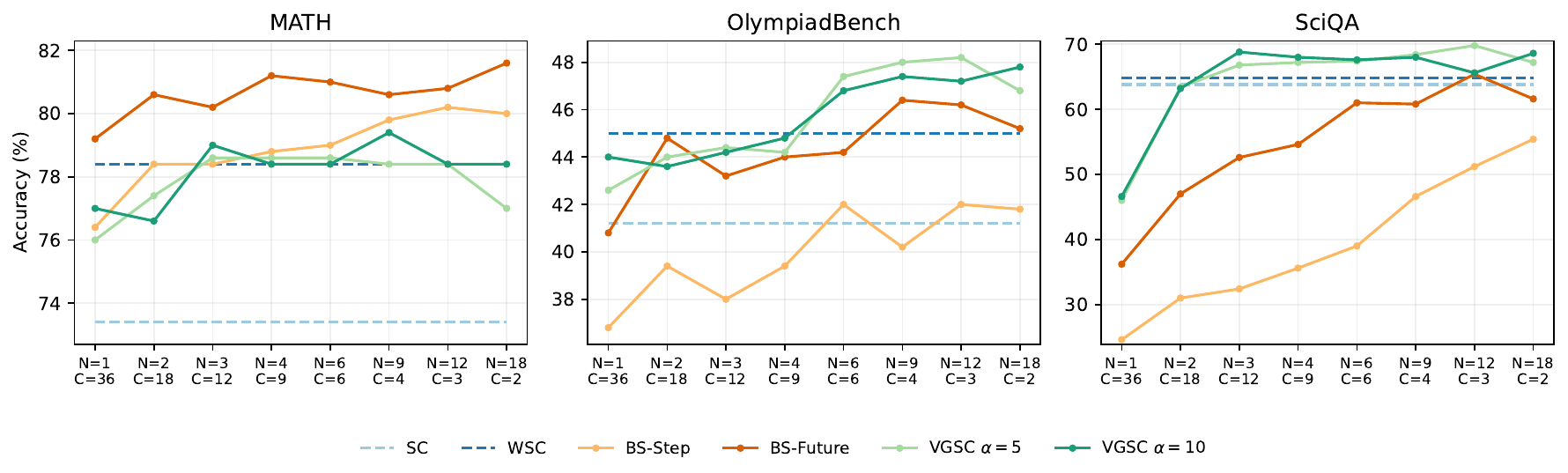}
    \caption{Allocation of search compute under a fixed budget. We fix $N \cdot C = 36$ and vary how the budget is split between the number of reasoning chains $N$ and candidates $C$.}
    \label{fig:ablation_fixed_nc}
  \end{figure*}

  Figure~\ref{fig:ablation_fixed_nc} fixes the total search budget to $N \cdot C = 36$ and varies how it is split between cross-trajectory exploration, controlled by $N$, and within-step verifier-guided resampling, controlled by $C$. For reference, SC and WSC use $36$ complete trajectories. VGSC remains competitive across almost all allocations, indicating that its gains do not depend on a narrowly tuned choice of $N$ and $C$. The best performance is typically achieved at intermediate allocations, around $N \in \{6,9\}$, suggesting that both sources of diversity are useful: too few trajectories limits final self-consistency, while too few candidates per step weakens verifier-guided steering. This supports the balanced $N=C$ setting used in the main experiments.

  \begin{table*}[h]
  \centering
  \small
  \caption{Accuracy $\uparrow$ across datasets for \qwen generator with \qwenprm verifier. Results are averaged over $5$ random seeds. The best result in each column is \textbf{bold} and the second-best result is \underline{underlined}. Product aggregation has overall the best performance.}
  \begin{tabular}{lc|ccccc|c}
    \toprule
    Method & \multirowcell{Verifier\\Aggregation} & MATH & \multirowcell{Olympiad\\Bench} & SciQA & CsQA & K\&K & \multirowcell{Average\\Accuracy}\\
\midrule
 \rowcolor{gray!20}
 \multicolumn{8}{c}{\textit{Post-hoc verifier aggregation}} \\
\midrule
WSC & product & 80.5 & 43.6 & 67.0 & 49.8 & 56.2 & 59.4 \\
WSC & min & 80.1 & 43.2 & 67.5 & 49.6 & \textbf{65.3} & 61.1 \\
WSC & mean & 77.5 & 39.3 & 64.0 & 48.0 & 55.7 & 56.9 \\
\midrule
DLE & product & 78.6 & 43.8 & 67.5 & 52.8 & 60.4 & 60.6 \\
DLE & min & 79.2 & 42.7 & 67.2 & 52.4 & 58.0 & 59.9\\
DLE & mean & 76.4 & 41.2 & 67.2 & 50.6 & 55.0 & 58.1 \\
\midrule
 \rowcolor{gray!20}
 \multicolumn{8}{c}{\textit{Verifier-guided search baselines}} \\
\midrule
BS-Step & product & 77.4 & 38.5 & 44.8 & 31.1 & 38.0 & 46.0 \\
BS-Step & min & 76.7 & 37.7 & 47.0 & 30.3 & 37.3 & 45.8 \\
BS-Step & mean & 77.5 & 38.2 & 46.3 & 31.0 & 40.7 & 46.7 \\
\midrule
BS+SC-Step & product & 78.4 & 41.1 & 62.1 & 46.7 & 56.6 & 57.0 \\
BS+SC-Step & min & 77.7 & 41.2 & 63.1 & 48.5 & 56.3 & 57.4 \\
BS+SC-Step & mean & 78.6 & 41.5 & 61.5 & 47.1 & 54.3 & 56.6 \\
\midrule
 \rowcolor{gray!20}
 \multicolumn{8}{c}{\textit{Verifier-guided self-consistency (ours)}} \\
\midrule
VGSC $\alpha=10$ & product & \textbf{81.1}& 45.0& \textbf{70.3}& \underline{54.5}& 64.5& \underline{63.1} \\
VGSC $\alpha=10$ & min & \underline{80.6}& \underline{45.2}& \underline{70.2}& 51.6& 61.0& 61.7 \\
VGSC $\alpha=10$ & mean & 79.4& \textbf{45.9}& \underline{70.2}& \textbf{56.0}& \underline{65.0}& \textbf{63.3} \\
    \bottomrule
  \end{tabular}
  \label{tab:ablation_aggregations}
\end{table*}


\newpage

\section{Analysis and Examples}
\label{app:analysis}
  In this section we provide a qualitative and quantitative analysis of the decoding strategies considered in the main paper.

\subsection{Qualitative Examples}
  We first present qualitative examples illustrating the characteristic failure modes of baseline methods as well as VGSC. We analyze representative generations from OlympiadBench and SciQA using \qwen and \llama generators with the \qwenprm verifier for each baseline.
  
\paragraph{Example 1: VGSC vs WSC.}
  A typical failure mode of WSC arises when all sampled trajectories are incorrect because the generator fails to produce a correct solution, even though the verifier can identify low-quality or erroneous intermediate steps. Since WSC applies verifier information only after complete trajectories have already been generated, it cannot use this signal to redirect decoding toward a more promising solution. In contrast, VGSC uses verifier guidance during decoding, which can steer generation away from erroneous continuations and toward a correct final answer. Figure~\ref{fig:analysis-wsc-failure} shows a representative example of this behavior.
  
\begin{figure*}[h]
\centering
\scriptsize
\fbox{
  \parbox{0.97\textwidth}{
    \textbf{Problem.} \\
    Let $EUCLID$ be a hexagon inscribed in a circle of radius $5$. Given that $EU = UC = LI = ID = 6$ and $CL = DE$, compute $CL$. \\
    \textbf{Gold answer:} $\boxed{14/5}$

    \vspace{0.8em}
    \begin{minipage}[t]{0.37\textwidth}
      \textbf{WSC trajectory (truncated).}
      \vspace{0.3em} \\
      \emph{Step 1 ($V_\phi =$ {\bf\color{darkgreen}0.956}):} \quad Since $EU = UC = LI = ID = 6$, these four equal chords subtend equal central angles; let each such angle be $\theta$.
      \vspace{0.0em} \\
      \emph{Step 2 ($V_\phi =$ {\bf\color{darkred}0.179}):} \quad The four equal angles account for half the circle, so $4\theta = 180^\circ$, hence $\theta = 45^\circ$.
      \vspace{1.2em} \\
      \emph{Step 3 ($V_\phi =$ {\bf\color{darkyellow}0.732}):} \quad Therefore, $CL = DE = 10\sin(22.5^\circ) = \boxed{5\sqrt{2-\sqrt{2}}}$. \color{darkred}{\textbf{\large{\ding{55}}}}
    \end{minipage}
    \hfill
    \begin{minipage}[t]{0.57\textwidth}
      \textbf{VGSC trajectory (truncated).}
      \vspace{0.8em} \\
      \emph{Step 1 ($V_\phi =$ {\bf\color{darkgreen}0.919}):} \quad Let $CL = DE = x$. Let each length-$6$ chord subtend central angle $\theta$, and let each chord of length $x$ subtend central angle $\phi$.
      \vspace{0.8em} \\
      \emph{Step 2 ($V_\phi =$ {\bf\color{darkgreen}0.959}):} \quad Then $4\theta + 2\phi = 360^\circ$, so $\phi = 180^\circ - 2\theta$.
      \vspace{0.8em} \\
      \emph{Step 3 ($V_\phi =$ {\bf\color{darkgreen}0.960}):} \quad From $6 = 10\sin(\theta/2)$, we get $\sin(\theta/2) = 3/5$ and $\cos(\theta/2)$ = $4/5$, hence $\cos(\theta) = 2(4/5)^2 - 1 = 7/25$.
      \vspace{0.8em} \\
      \emph{Step 4 ($V_\phi =$ {\bf\color{darkgreen}0.954}):} \quad So, $x = 10\sin(\phi/2) = 10\cos(\theta) = 10 \cdot \frac{7}{25} = \boxed{14/5}$. \color{darkgreen}{\textbf{\large{\ding{51}}}}
    \end{minipage}
  }
}
\caption{Truncated example from OlympiadBench where WSC fails but VGSC succeeds. WSC samples complete trajectories from the base generator and applies verifier information only after generation, so it cannot redirect decoding once it enters an incorrect line of reasoning. In this example, all WSC trajectories contain low-scored incorrect steps and end in incorrect answers. VGSC instead uses verifier guidance during decoding and is steered toward a correct solution.}
\label{fig:analysis-wsc-failure}
\vspace{-0.5em}
\end{figure*}
  
\paragraph{Example 2: VGSC vs BS-Step.}
  A common failure mode of BS-Step is that it overemphasizes local correctness, leading the search to prefer steps that are individually plausible but do not make meaningful progress toward the final solution. As a result, beam search can become trapped in locally high-scoring but unproductive reasoning paths. VGSC mitigates this issue by incorporating future-outcome guidance in addition to step-level verification, which favors trajectories that are both locally sound and globally promising. Figure~\ref{fig:analysis-bs-step-failure} illustrates this pattern.

\begin{figure*}[!h]
\centering
\scriptsize
\fbox{
  \parbox{0.97\textwidth}{
    \textbf{Problem.} \\
    For a positive integer $a$, define a sequence of integers $x_1, x_2, \ldots$ by letting $x_1 = a$ and $x_{n+1} = 2x_n + 1$ for $n \geq 1$. Let $y_n = 2^{x_n} - 1$. Determine the largest possible $k$ such that, for some positive integer $a$, the numbers $y_1, \ldots, y_k$ are all prime. \\
    \textbf{Gold answer:} $\boxed{2}$

    \vspace{0.8em}
    \begin{minipage}[t]{0.42\textwidth}
      \textbf{BS-Step trajectory (truncated).}
      \vspace{0.8em} \\
      \emph{Step 1 ($V_\phi =$ {\bf\color{darkgreen}0.999}):} \quad We analyze the sequences $x_n$ and $y_n$, where $x_1 = a$, $x_{n+1} = 2x_n + 1$, and $y_n = 2^{x_n} - 1$.
      \vspace{0.4em} \\
      \emph{Step 2 ($V_\phi =$ {\bf\color{darkgreen}1.000}):} \quad The first terms are $x_2 = 2a + 1$ and $x_3 = 4a + 3$
      \vspace{0.4em} \\
      \emph{Step 3 ($V_\phi =$ {\bf\color{darkgreen}1.000}):} \quad Next, $x_4 = 8a + 7$ and $x_5 = 16a + 15$
      \vspace{0.6em} \\
      \emph{Step 4 ($V_\phi =$ {\bf\color{darkgreen}1.000}):} \quad Next, $x_6 = 32a + 31$ and $x_7 = 64a + 63$
      \vspace{-0.6em} \\
      \emph{Step 5 ($V_\phi =$ {\bf\color{darkgreen}1.000}):} \quad Next, $x_8 = 128a + 127$ and $x_9 = 256a + 255$
      \vspace{0.0em} \\
      \emph{\dots}
      \vspace{0.2em} \\
      \emph{Step 11 ($V_\phi =$ {\bf\color{darkgreen}1.000}):} \quad $x_{20} = 524288a + 524287$ and $x_{21} = 1048576a + 1048575$ \hfill \emph{\dots} \color{darkred}{\textbf{\large{\ding{55}}}}
    \end{minipage}
    \hfill
    \begin{minipage}[t]{0.52\textwidth}
      \textbf{VGSC trajectory (truncated).}
      \vspace{0.8em} \\
      \emph{Step 1 ($V_\phi =$ {\bf\color{darkgreen}0.903}):} \quad If $2^m - 1$ is prime, then $m$ must be prime. Therefore, if $y_1$ and $y_2$ are both prime, then $a$ and $2a+1$ must both be prime.
      \vspace{0.8em} \\
      \emph{Step 2 ($V_\phi =$ {\bf\color{darkgreen}0.944}):} \quad Taking $a = 2$ gives $x_1 = 2$, $x_2 = 5$, and $x_3 = 11$. Hence $y_1 = 2^2 - 1 = 3$ and $y_2 = 2^5 - 1 = 31$ are prime, while $y_3 = 2^{11} - 1 = 2047 = 23 \cdot 89$ is composite. Thus $k \geq 2$.
      \vspace{0.8em} \\
      \emph{Step 3 ($V_\phi =$ {\bf\color{darkgreen}0.939}):} \quad It remains to show that $k$ cannot exceed $2$. If $a$ is an odd prime and $y_2$ is prime, then $q = 2a+1$ is an odd prime with $q \equiv 3 \pmod{4}$. Let $p = x_3 = 2q + 1$. If $p$ is composite, then $y_3 = 2^p - 1$ is composite.
      \vspace{0.8em} \\
      \emph{Step 4 ($V_\phi =$ {\bf\color{darkgreen}0.989}):} \quad If $p$ is prime, then $p \equiv 7 \pmod{8}$, so $2^q \equiv 1 \pmod{p}$. Hence $p \mid (2^q - 1) = y_2$, contradicting the primality of $y_2$. Therefore $y_3$ is never prime, and the largest possible value is $\boxed{2}$. \color{darkgreen}{\textbf{\large{\ding{51}}}}
    \end{minipage}
  }
}
\caption{Truncated example where BS-Step fails but VGSC succeeds. BS-Step ranks partial trajectories only by local step-verifier scores $V_\phi$, so it keeps selecting steps that are individually correct but do not move the solution forward. In this example, beam search repeatedly expands the recurrence instead of reaching the key primality and divisibility argument. VGSC adds future-outcome guidance, which favors trajectories that make progress toward a complete proof and therefore reaches the correct answer.}
\label{fig:analysis-bs-step-failure}
\end{figure*}

\paragraph{Example 3: VGSC vs BS-Future.}
  A common failure mode of BS-Future is over-concentration on a narrow region of the search space when the future-outcome predictor is imperfect. In such cases, beam search may continue expanding a flawed partial trajectory because its future score is somewhat inflated, even when that trajectory is ultimately unproductive. VGSC instead combines verifier guidance with generator sampling, which preserves exploration and reduces sensitivity to this kind of predictor noise. Figure~\ref{fig:analysis-bs-future-failure} provides a representative example.

\begin{figure*}[!h]
\centering
\scriptsize
\fbox{
  \parbox{0.97\textwidth}{
    \textbf{Problem.} \\
    Hannah flips two fair coins, while Otto flips $T=5$ fair coins. Let $p$ be the probability that the number of heads on Hannah's coins is greater than the number of heads on Otto's coins. If $p = q/r$, where $q$ and $r$ are relatively prime positive integers, compute $q+r$. \\
    \textbf{Gold answer:} $\boxed{17}$

    \vspace{0.8em}
    \begin{minipage}[t]{0.45\textwidth}
      \textbf{BS-Future trajectory (truncated).}
      \vspace{0.8em} \\
      \emph{Step 1 ($\widehat{Q} =$ {\bf\color{darkyellow}0.620}):} \quad Let $H$ be the number of heads from Hannah's two coins, $P(\text{H=0})=1/4$ $P(\text{H=1})=1/2$, and $P(\text{H=2})=1/4$.
      \vspace{0.3em} \\
      \emph{Step 2 ($\widehat{Q} =$ {\bf\color{darkyellow}0.686}):} \quad $O$ - number of heads from Otto's coins, $P(O=k)=\binom{5}{k}/32$
      \vspace{0.3em} \\
      \emph{Step 3 ($\widehat{Q} =$ {\bf\color{darkgreen}0.828}):} \quad There are 3 cases when H>O: H=1, O=0; H=2, O=0 or H=2, O=1
      \vspace{0.3em} \\
      \emph{Step 4 ($\widehat{Q} =$ {\bf\color{darkgreen}0.851}):} \quad Case 1: P(H=1, O=0) $= \frac{1}{2} \cdot \frac{1}{32} = \frac{1}{64}$
      \vspace{0.2em} \\
      \emph{Step 5 ($\widehat{Q} =$ {\bf\color{darkgreen}0.901}):} \quad Case 2: P(H=1, O=0) $= \frac{1}{2} \cdot \frac{1}{32} = \frac{1}{64}$
      \vspace{0.2em} \\
      \emph{Step 6 ($\widehat{Q} =$ {\bf\color{darkgreen}0.901}):} \quad Case 3: P(H=1, O=0) $= \frac{1}{2} \cdot \frac{1}{32} = \frac{1}{64}$
      \vspace{0.2em} \\
      \emph{Step 7 ($\widehat{Q} =$ {\bf\color{darkgreen}0.902}):} \quad Case 4: P(H=1, O=0) $= \frac{1}{2} \cdot \frac{1}{32} = \frac{1}{64}$
      \vspace{0.2em} \hfill \emph{\dots} \color{darkred}{\textbf{\large{\ding{55}}}}
    \end{minipage}
    \hfill
    \begin{minipage}[t]{0.49\textwidth}
      \textbf{VGSC trajectory (truncated).}
      \vspace{0.8em} \\
      \emph{Step 1 ($\widehat{Q} =$ {\bf\color{darkyellow}0.680}):} \quad Let $H$ be the number of heads from Hannah's two coins and $O$ the number of heads from Otto's five coins. Then $P(\text{H=0})=1/4$, $P(\text{H=1})=1/2$, and $P(\text{H=2})=1/4$, while $P(\text{O=k})=\binom{5}{k}/32$.
      \vspace{0.5em} \\
      \emph{Step 2 ($\widehat{Q} =$ {\bf\color{darkyellow}0.655}):} \quad The event H>O can occur only in two cases: $(\text{H=1},\text{O=0})$ or $(\text{H=2},\text{O$\leq$1})$.
      \vspace{0.8em} \\
      \emph{Step 3 ($\widehat{Q} =$ {\bf\color{darkgreen}0.760}):} \quad Therefore, P(H>O) = P(H=1)P(O=0) + P(H=2)P(O<2).
      \vspace{0.2em} \\
      \emph{Step 4 ($\widehat{Q} =$ {\bf\color{darkgreen}0.935}):} \quad Compute
      P(H=1) P(O=0) $= \frac{1}{2}\cdot\frac{1}{32} = \frac{1}{64}$,
      and
      P(H=2) P(O<2) $= \frac{1}{4}\left(\frac{1}{32}+\frac{5}{32}\right)=\frac{3}{64}$.
      Hence
      $p=\frac{1}{64}+\frac{3}{64}=\frac{1}{16}$, so $q+r=1+16=\boxed{17}$. \color{darkgreen}{\textbf{\large{\ding{51}}}}
    \end{minipage}
  }
}
\caption{Truncated example from OlympiadBench where BS-Future fails but VGSC succeeds. BS-Future ranks partial trajectories only by the future-outcome predictor $\widehat{Q}$, so moderately inflated future scores can keep the beam focused on a flawed branch. In this example, beam search commits to an incorrect case analysis and keeps expanding it without reaching other cases. VGSC preserves exploration due to independent sampling from verifier-tilted distribution and reaches the correct probability computation in majority of samples.}
\label{fig:analysis-bs-future-failure}
\end{figure*}

\paragraph{Example 4: VGSC failure.}
  VGSC can also fail when the verifier is poorly aligned with the task. In this regime, the verifier may systematically prefer incorrect trajectories, so that guiding decoding toward the verifier-tilted distribution becomes harmful rather than helpful. This can make verifier-guided methods underperform simpler generator-only aggregation methods such as self-consistency. Figure~\ref{fig:analysis-vgsc-failure} shows a representative failure case of this kind.
  
\begin{figure*}[!h]
\centering
\scriptsize
\fbox{
  \parbox{0.97\textwidth}{
    \textbf{Problem.} \\
    Would you find the word \emph{pink} on a dictionary page with the following guide words? \emph{parrot -- property} \quad Options: $(1)$ yes \; $(2)$ no \hfill \\
    \textbf{Gold answer:} $\boxed{1}$

    \vspace{0.8em}
    \begin{minipage}[t]{0.60\textwidth}
      \textbf{SC/WSC samples (truncated).}
      \vspace{0.5em} \\
      \emph{Sample 1 ($V_\phi =$ {\bf\color{darkred}0.001}):} \quad To decide whether \emph{pink} would be found on a dictionary page with guide words \emph{parrot} and \emph{property}, we compare the words. These guide words are not directly related to \emph{pink}. Therefore, \emph{pink} would not be found on the page. Thus, the answer is $\boxed{1}$ {\color{darkgreen}{\textbf{\large{\ding{51}}}}}
      \vspace{0.5em} \\
      \emph{Sample 2 ($V_\phi =$ {\bf\color{darkred}0.074}):} \quad The guide words are \emph{parrot} and \emph{property}. Although these words are not directly related to \emph{pink}, the word \emph{pink} is commonly found in a standard dictionary. Therefore, it is reasonable to conclude that \emph{pink} is present. Hence, the answer is $\boxed{1}$ {\color{darkgreen}{\textbf{\large{\ding{51}}}}}
      \vspace{0.5em} \\
      \emph{Sample 3 ($V_\phi =$ {\bf\color{darkyellow}0.416}):} \quad The guide words are \emph{parrot} and \emph{property}. The word \emph{pink} is a color and is not directly related to either guide word. Therefore, based on the guide words, \emph{pink} would not be found on the dictionary page. Thus, the answer is $\boxed{2}$  {\color{darkred}{\textbf{\large{\ding{55}}}}}
      \vspace{0.5em} \\
      \textbf{SC vote:} $\boxed{1}$  {\color{darkgreen}{\textbf{\large{\ding{51}}}}} \hfill
      \textbf{WSC vote:} $\boxed{2}$ {\color{darkred}{\textbf{\large{\ding{55}}}}} \hfill \hfill
    \end{minipage}
    \hfill
    \begin{minipage}[t]{0.35\textwidth}
      \textbf{VGSC trajectory (truncated).}
      \vspace{0.5em} \\
      \emph{Step 1 ($V_\phi =$ {\bf\color{darkgreen}0.722}):} \quad To determine whether the word \emph{pink} would be found on a dictionary page with the guide words \emph{parrot} and \emph{property}, we consider the context and meaning of the guide words.
      \vspace{0.5em} \\
      \emph{Step 2 ($V_\phi =$ {\bf\color{darkgreen}0.840}):} \quad Identify the guide words: \emph{parrot} and \emph{property}.
      \vspace{0.5em} \\
      \emph{Step 3 ($V_\phi =$ {\bf\color{darkyellow}0.591}):} \quad Neither of these words has any direct relationship to \emph{pink}, which is a color.
      \vspace{0.5em} \\
      \emph{Step 4 ($V_\phi =$ {\bf\color{darkyellow}0.479}):} \quad Therefore, the word \emph{pink} does not have an explicit connection to these guide words.
      \vspace{0.5em} \\
      \emph{Step 5 ($V_\phi =$ {\bf\color{darkgreen}0.717}):} \quad Thus, the final answer is $\boxed{2}$. \color{darkred}{\textbf{\large{\ding{55}}}}
    \end{minipage}
  }
}
\caption{VGSC failure case from SciQA. In this example, plain self-consistency over base-model samples returns the correct answer $\boxed{1}$, but the verifier is poorly aligned with the task and assigns higher scores to chains that confuse dictionary guide words with semantic relatedness. Consequently, WSC selects the incorrect answer $\boxed{2}$, and VGSC is pulled toward the same incorrect region under verifier-guided decoding. This shows that misaligned verifier signals can make verifier-tilted sampling worse than using the base generator alone.}
\label{fig:analysis-vgsc-failure}
\end{figure*}

\begin{table*}[!ht]
\centering
\small
\caption{Fraction of VGSC decoding steps that reach fallback for \qwen{} + PRM.}
\begin{tabular}{l|ccccc|c}
  \toprule
  Dataset & MATH & \multirowcell{Olympiad\\Bench} & SciQA & CsQA & K\&K & Average \\
  \midrule
    $\alpha=1$  & 10.76\% & 43.66\% & 57.15\% & 63.54\% & 56.23\% & 46.27\% \\
    $\alpha=5$  & 16.88\% & 73.97\% & 72.00\% & 78.26\% & 68.01\% & 61.82\% \\
    $\alpha=10$ & 20.45\% & 80.79\% & 76.34\% & 82.59\% & 73.21\% & 66.68\% \\
    $\alpha=20$ & 26.12\% & 85.49\% & 80.27\% & 85.59\% & 77.48\% & 70.99\% \\
  \bottomrule
\end{tabular}
\label{tab:fallback_statistics}
\end{table*}

\subsection{Coverage and chain diversity}
  To better understand the behavior of different methods, we complement the main accuracy results with simple statistics of candidate-set quality and whole-chain diversity. For each example, let $\{\yv^{(1)}, \dots, \yv^{(N)}\}$ denote the completed reasoning chains generated by a method. We report:
  \begin{itemize}[leftmargin=12pt, itemsep=0.2pt, topsep=0.4pt, parsep=0pt]
    \item \text{Accuracy}: final exact-match accuracy after aggregation.
    \item \text{Pass@N}: whether at least one of the $N$ generated chains is correct.
    \item \text{Diversity}: one minus the mean pairwise ROUGE-L similarity \citep{lin2004rouge} between completed chains from the same example:
    \begin{align}
      &\text{Diversity}(x) =
    1 - \frac{2}{N(N-1)} \sum_{i<j} \mathrm{ROUGE\mbox{-}L}\!\left(\yv^{(i)}, \yv^{(j)}\right).
    \end{align}
  \end{itemize}
  Higher values of \text{Pass@N} and \text{Diversity} indicate better candidate coverage and greater diversity.
  \begin{table*}[!t]
    \centering
    \small
    \caption{Accuracy, average pass@N and whole-chain diversity statistics on random 100-example subsets of MATH and OlympiadBench using \qwen with the \qwenprm verifier.}
    \begin{tabular}{l|ccc|ccc}
      \toprule
      \multirow{2}{*}{Method} & \multicolumn{3}{c|}{MATH} & \multicolumn{3}{c}{OlympiadBench} \\
       & Accuracy $\uparrow$ & Pass@N $\uparrow$ & Diversity $\uparrow$ & Accuracy $\uparrow$ & Pass@N $\uparrow$ & Diversity $\uparrow$ \\
        \midrule
         \rowcolor{gray!20}
         \multicolumn{7}{c}{\textit{No-verifier baselines}} \\
        \midrule
      SC & 76 & 80 & \textbf{0.487} & 40 & 46 & \textbf{0.551} \\
        \midrule
         \rowcolor{gray!20}
         \multicolumn{7}{c}{\textit{Post-hoc verifier aggregation}} \\
        \midrule
      WSC & 78 & 80 & \textbf{0.487} & \underline{45} & 46 & \textbf{0.551} \\
        \midrule
         \rowcolor{gray!20}
         \multicolumn{7}{c}{\textit{Verifier-guided search baselines}} \\
        \midrule
      BS-Step & 77 & 78 & 0.147 & 38 & 43 & 0.177 \\
      BS-Future & {\textbf{80}}& {82}& {0.140}& {43}& {46}& {0.168} \\
        \midrule
         \rowcolor{gray!20}
         \multicolumn{7}{c}{\textit{Verifier-guided self-consistency (ours)}} \\
        \midrule
      VGSC $\alpha = 5$ & {\textbf{80}}& {\textbf{85}}& {\underline{0.284}}& {\textbf{47}}& {\textbf{52}}& {\underline{0.436}} \\
      VGSC $\alpha = 10$ & {\underline{79}}& {\underline{84}}& {0.277}& {\underline{45}}& {\underline{49}}& {0.429} \\
      \bottomrule
    \end{tabular}
    \label{tab:analysis_diversity}
\end{table*}

  Table~\ref{tab:analysis_diversity} reports the average values of the above metrics for the two datasets, highlighting the key differences among methods. SC and WSC often generate diverse reasoning chains; however, because they do not incorporate verifier guidance during generation, they tend to achieve lower accuracy. By contrast, BS-Step applies verifier guidance throughout decoding, which improves accuracy but often reduces diversity, as the search concentrates on a narrow set of trajectories. VGSC offers a middle ground: it generates multiple diverse chains while still leveraging verifier guidance during decoding, achieving strong Accuracy and \text{Pass@N} without suffering the same degree of diversity collapse as beam search.

\subsection{Fallback Statistics}
  \label{app:fallback_statistics}

  Theorem~\ref{theorem:vgsc} shows that finite-cap bias occurs only when all $C$ proposed next steps are rejected and VGSC falls back to restricted selection. We therefore measure how often this fallback occurs under the default cap $C=5$. Table~\ref{tab:fallback_statistics} reports fallback rates for different verifier-trust exponents $\alpha$. As expected, fallback becomes more frequent as $\alpha$ increases, since larger values make $V^\alpha \widehat{Q}^\alpha$ more selective. Nevertheless, even at $\alpha=20$, fewer than $70\%$ of decoding steps reach fallback on average. On MATH, fallback is even rarer, occurring in only $10\%$--$26\%$ of steps across $\alpha$ variants.

\section{Theoretical Results}
\label{app:proofs}
  This section proves Proposition~\ref{prop:local_tilted_conditional} and Theorem~\ref{theorem:vgsc}. We also derive a trajectory-level implication of the one-step bound and analyze the stability of verifier-tilted distributions under imperfect verifier scores.

\subsection{Proof of Proposition~\ref{prop:local_tilted_conditional}}
\label{app:proposition_proof}
  By definition,
  \begin{equation*}
    P^V_\alpha(\yv \mid \xv)
    \propto
    P(\yv \mid \xv)V(\yv \mid \xv)^\alpha.
  \end{equation*}
  For a fixed prefix $\yv_{<t}$ and candidate step $y$, the conditional probability under $P^V_\alpha$ is obtained by summing over all suffixes:
  \begin{equation*}
    P^V_\alpha(y_t=y \mid \xv,\yv_{<t})
    \propto
    \sum_{\yv_{>t}} P(\yv \mid \xv)V(\yv \mid \xv)^\alpha.
  \end{equation*}
  Using the chain rule for the generator and the stepwise verifier factorization, and absorbing all prefix-dependent terms into the proportionality constant, we get
  \begin{align}
    P^V_\alpha(y_t=y \mid \xv,\yv_{<t})
    &\propto
    P(y \mid \xv,\yv_{<t})
    V(y \mid \xv,\yv_{<t})^\alpha \cdot
    \sum_{\yv_{>t}}
    P(\yv_{>t} \mid \xv,\yv_{<t}\circ y) V(\yv_{>t} \mid \xv,\yv_{<t}\circ y)^\alpha = \\ &=
    P(y \mid \xv,\yv_{<t})
    V(y \mid \xv,\yv_{<t})^\alpha
    Q_\alpha(\yv_{<t}\circ y,\xv). \nonumber
  \end{align}
  This proves Proposition~\ref{prop:local_tilted_conditional}. \hfill $\square$

\subsection{Proof of Theorem~\ref{theorem:vgsc}} 
\label{app:theorem_proof}
  Fix a prefix $\yv_{<t}$ and write
  \begin{equation}
    \begin{aligned}
      p_t(y) &:= P(y \mid \xv,\yv_{<t}), \\
      w_t(y) &:= W_\alpha(y \mid \xv,\yv_{<t}), \\
      Z_t &:= \mathbb{E}_{Y\sim p_t}[w_t(Y)].
    \end{aligned}
  \end{equation}
  By the exact-guidance assumption,
  \begin{equation}
    \begin{aligned}
      q_t(y)=
      P^V_\alpha(y_t=y \mid \xv,\yv_{<t}) = \frac{p_t(y)w_t(y)}{Z_t}.
    \end{aligned}
  \end{equation}
  If $Z_t=1$, fallback has probability zero and the result is immediate. Thus assume
  $0<Z_t<1$.

  Let $F_C$ be the event that all $C$ proposals are rejected. A single proposal is accepted with
  marginal probability $Z_t$, so
  \begin{equation}
    \Pr(F_C \mid \xv,\yv_{<t})=(1-Z_t)^C
    \le
    \exp(-CZ_t).
  \end{equation}

  Conditioned on not reaching fallback, the output is exactly distributed as $q_t$. Indeed, the
  probability that the first accepted proposal has value $y$ and occurs at attempt $j$ is
  $(1-Z_t)^{j-1}p_t(y)w_t(y)$. Therefore
  \begin{align}
    &\Pr(\tilde y_t=y, F_C^c \mid \xv,\yv_{<t}) =
    \sum_{j=1}^C
    (1-Z_t)^{j-1}p_t(y)w_t(y) =
    \frac{1-(1-Z_t)^C}{Z_t}
    p_t(y)w_t(y).
  \end{align}
  Dividing by $\Pr(F_C^c \mid \xv,\yv_{<t})=1-(1-Z_t)^C$ gives
  \begin{equation}
    \Pr(\tilde y_t=y \mid F_C^c,\xv,\yv_{<t})
    =
    q_t(y).
  \end{equation}

  It remains to bound the fallback contribution. Conditioned on rejection, each cached proposal is
  distributed as
  \begin{equation}
    \bar{p}_t(y)
    =
    \frac{p_t(y)(1-w_t(y))}{1-Z_t}.
  \end{equation}
  The fallback uses odds weights
  \begin{equation}
    \rho_t(y):=\frac{w_t(y)}{1-w_t(y)}.
  \end{equation}
  Hence
  \begin{equation}
    \bar{p}_t(y)\rho_t(y)
    =
    \frac{p_t(y)w_t(y)}{1-Z_t}
    \propto
    q_t(y),
  \end{equation}
  so the fallback step is a sampling-importance-resampling (SIR) approximation to $q_t$ from samples
  drawn from $\bar{p}_t$.

  Let $r_{C,t,\alpha}$ denote the marginal distribution of the fallback sample. Define
  \begin{equation}
    \begin{aligned}
      \eta_t
      &:=
      \mathbb{E}_{Y\sim \bar{p}_t}[\rho_t(Y)]
      =
      \frac{Z_t}{1-Z_t}, \\
      \Gamma_t
      &:=
      \mathbb{E}_{Y\sim \bar{p}_t}
      \left[
        \left(\frac{\rho_t(Y)}{\eta_t}\right)^2
      \right].
    \end{aligned}
  \end{equation}
  The boundedness assumption $0<\epsilon\le w_t(y)\le 1-\epsilon$ implies
  $\epsilon\le Z_t\le 1-\epsilon$ and
  \begin{equation}
    \begin{aligned}
      \Gamma_t
      &=
      \frac{1-Z_t}{Z_t^2}
      \sum_y p_t(y)
      \frac{w_t(y)^2}{1-w_t(y)} \le
      \frac{(1-\epsilon)^3}{\epsilon^3}
      <\infty.
    \end{aligned}
  \end{equation}
  By the standard non-asymptotic bias bound for self-normalized importance sampling over bounded
  test functions \citep[Theorem~2.1]{Agapiou2015ImportanceSI}, and since the marginal law of the
  SIR fallback sample is the expectation of the self-normalized weighted empirical measure, there
  exists a constant $B_t<\infty$, independent of $C$, such that
  \begin{equation}
    d_{\mathrm{TV}}(r_{C,t,\alpha},q_t)
    \le
    \frac{B_t}{C}.
  \end{equation}

  Finally, the bounded VGSC transition decomposes as
  \begin{equation}
    \begin{aligned}
      \mathcal{K}_{C,t,\alpha}(\cdot \mid \xv,\yv_{<t})
      &=
      \left(1-(1-Z_t)^C\right)q_t(\cdot) +
      (1-Z_t)^C r_{C,t,\alpha}(\cdot).
    \end{aligned}
  \end{equation}
  Therefore,
  \begin{align}
    &d_{\mathrm{TV}}\!\left(
      \mathcal{K}_{C,t,\alpha}(\cdot \mid \xv,\yv_{<t}),
      q_t(\cdot)
    \right) =
    (1-Z_t)^C d_{\mathrm{TV}}(r_{C,t,\alpha},q_t) \le
    B_t\frac{(1-Z_t)^C}{C}
    \le
    B_t\frac{\exp(-CZ_t)}{C}.
  \end{align}
  \hfill $\square$

\subsection{Trajectory-level implication.}
\label{app:proof_trajectory_level}
  The one-step bound also gives a simple bound on complete trajectories. Assume that all generated
  trajectories have length at most $T$. Let $q_t(\cdot \mid \xv,\yv_{<t})$ denote the ideal local
  conditional at step $t$, and let $\mathcal{K}_{C,t,\alpha}(\cdot \mid \xv,\yv_{<t})$ denote the bounded
  VGSC transition. Define
  \begin{equation}
    \begin{aligned}
      &\varepsilon_t(\xv,\yv_{<t}) := d_{\mathrm{TV}}\!\left(
        \mathcal{K}_{C,t,\alpha}(\cdot \mid \xv,\yv_{<t}),
        q_t(\cdot \mid \xv,\yv_{<t})
      \right).
    \end{aligned}
  \end{equation}
  By Theorem~\ref{theorem:vgsc}, for every reachable prefix,
  \begin{equation*}
    \varepsilon_t(\xv,\yv_{<t}) \le B_t(\xv,\yv_{<t}) \cdot \frac{\exp(-CZ_t(\xv,\yv_{<t}))}{C}.
  \end{equation*}

  Let $\widehat P_C(\cdot \mid \xv)$ be the trajectory distribution induced by bounded VGSC, and
  let $P^\star(\cdot \mid \xv)$ be the trajectory distribution obtained by composing the ideal local
  conditionals $q_t$. A standard step-by-step maximal coupling gives
  \begin{equation*}
      d_{\mathrm{TV}}\!\left(
        \widehat P_C(\cdot \mid \xv),
        P^\star(\cdot \mid \xv)
      \right)
      \le
      \sum_{t=1}^T
      \sup_{\yv_{<t}}
      \varepsilon_t(\xv,\yv_{<t}).
  \end{equation*}
  Therefore,
  \begin{align}
      &d_{\mathrm{TV}}\!\left(
        \widehat P_C(\cdot \mid \xv),
        P^\star(\cdot \mid \xv)
      \right) \le
      \sum_{t=1}^T
      \sup_{\yv_{<t}}
      B_t(\xv,\yv_{<t}) \cdot
      \frac{\exp(-CZ_t(\xv,\yv_{<t}))}{C}. \nonumber
  \end{align}
  In particular, if $Z_t(\xv,\yv_{<t})\ge z_{\min}>0$ and
  $B_t(\xv,\yv_{<t})\le B<\infty$ for all reachable prefixes, then
  \begin{equation*}
      d_{\mathrm{TV}}\!\left(
        \widehat P_C(\cdot \mid \xv),
        P^\star(\cdot \mid \xv)
      \right)
      \le
      TB\frac{\exp(-Cz_{\min})}{C}.
  \end{equation*}
  When the unclipped exact weights are used, $P^\star=P^V_\alpha$. When clipped weights are used,
  $P^\star$ is the corresponding clipped verifier-guided trajectory distribution.

\subsection{Stability under imperfect verifier scores.}
\label{app:proof_imperfect_models}
  The preceding results assume exact local guidance weights. We now note that the verifier-tilted target is stable to small verifier errors in log space. Let $V^\star$ denote an ideal verifier and let $\widehat V$ denote an imperfect verifier. Define
  \begin{equation}
    \begin{aligned}
      P^{V^\star}_\alpha(\yv \mid \xv)
      \propto
      P(\yv \mid \xv)
      V^\star(\yv \mid \xv)^\alpha,
      P^{\widehat V}_\alpha(\yv \mid \xv)
      \propto
      P(\yv \mid \xv)
      \widehat V(\yv \mid \xv)^\alpha.
    \end{aligned}
  \end{equation}
  Suppose that, for all trajectories $\yv$ with $P(\yv\mid\xv)>0$,
  \begin{equation}
    \begin{aligned}
      \left|
        \log \widehat V(\yv \mid \xv)
        -
        \log V^\star(\yv \mid \xv)
      \right| \le
      \delta .
    \end{aligned}
  \end{equation}
  Then
  \begin{equation}
    \begin{aligned}
      d_{\mathrm{TV}}\!\left(
        P^{\widehat V}_\alpha(\cdot \mid \xv),
        P^{V^\star}_\alpha(\cdot \mid \xv)
      \right)
      \le
      \tanh\!\left(\frac{\alpha\delta}{2}\right)
      \le
      \frac{\alpha\delta}{2}.
    \end{aligned}
  \end{equation}

  Indeed, writing
  \begin{equation}
    \begin{aligned}
      h(\yv)
      &:=
      \left(
        \frac{\widehat V(\yv \mid \xv)}
        {V^\star(\yv \mid \xv)}
      \right)^\alpha,
    \end{aligned}
  \end{equation}
  the log-error assumption implies $h(\yv)\in[e^{-\alpha\delta},e^{\alpha\delta}]$. The density ratio between the two tilted distributions is proportional to $h$, and a standard bounded-ratio argument gives the stated total-variation bound.

  This bound highlights the role of $\alpha$: stronger verifier guidance can improve search when the verifier is informative, but it also amplifies verifier errors linearly for small $\delta$. This supports treating $\alpha$ as a verifier-trust parameter rather than always making it large.

\section{Experimental Details}
\label{app:setup}

\subsection{Datasets, Prompts, and Examples}
\label{app:datasets_prompts_examples}

  \begin{table*}[t]
  \centering
  \small
  \setlength{\tabcolsep}{5pt}
  \caption{Prompt templates used for each benchmark group. The semantic content of the prompts is kept fixed across generator families; only the model-specific chat-template packaging is changed when needed.}
  \begin{tabularx}{\textwidth}{c c X c}
  \toprule
  Benchmark(s) & Task format & Prompt template & Expected final answer \\
  \midrule
  \tcell{MATH500,\\OlympiadBench}
  & Open-ended
  & Please reason step by step, and put your final answer within \texttt{\textbackslash boxed\{\}}.
  & \texttt{\textbackslash boxed\{<answer>\}} \\
  \midrule
  \tcell{ScienceQA,\\CommonsenseQA}
  & Multiple choice
  & Solve the problem carefully. Show your reasoning in short numbered steps. Then on the last line, output only the positive integer of the answer choice in exactly this format: \texttt{\textbackslash boxed\{(1|2|3|\dots)\}}.
  & \tcell{\texttt{\textbackslash boxed\{k\}},\\$k \in \{1,2,3,\dots\}$} \\
  \midrule
  Knights\&Knaves
  & \tcell{Multiple choice\\(binary)}
  & Solve the problem carefully. Show your reasoning in short numbered steps. Then on the last line, output only the positive integer of the answer choice in exactly this format: \texttt{\textbackslash boxed\{(1|2)\}}.
  & \tcell{\texttt{\textbackslash boxed\{k\}},\\$k \in \{1,2\}$} \\
  \bottomrule
  \end{tabularx}
  \label{tab:prompt_templates}
\end{table*}
  \begin{table*}[!h]
  \centering
  \small
  \setlength{\tabcolsep}{5pt}
  \caption{Examples from the five evaluation datasets using the exact dataset paths and splits from our experiments. For multiple-choice datasets, we show the full prompted question including answer choices, since the model is instructed to return the boxed index of the selected option.}
  \begin{tabularx}{\textwidth}{l X l}
  \toprule
  Dataset & Example input question & Gold answer \\
  \midrule
  MATH500 &
  Convert the point $(0,3)$ in rectangular coordinates to polar coordinates. Enter your answer in the form $(r,\theta)$, where $r > 0$ and $0 \le \theta < 2\pi$.
  &
  $\left(3,\frac{\pi}{2}\right)$ \\
  \midrule
  OlympiadBench &
  Xenia and Sergey play the following game. Xenia thinks of a positive integer $N$ not exceeding 5000. Then she fixes 20 distinct positive integers $a_{1}, a_{2}, \ldots, a_{20}$ such that, for each $k=1,2,\ldots,20$, the numbers $N$ and $a_{k}$ are congruent modulo $k$. By a move, Sergey tells Xenia a set $S$ of positive integers not exceeding 20, and she tells him back the set $\{a_k : k \in S\}$ without spelling out which number corresponds to which index. How many moves does Sergey need to determine for sure the number Xenia thought of?
  &
  $2$ \\
  \midrule
  ScienceQA &
  Which figure of speech is used in this text?

  Sing, O goddess, the anger of Achilles son of Peleus, that brought countless ills upon the Achaeans. --- Homer, \emph{The Iliad}

  Options:
  
  (1) chiasmus \qquad (2) apostrophe & 2 \\
  \midrule
  CommonsenseQA &
  A revolving door is convenient for two direction travel, but it also serves as a security measure at a what?

  Choices: 
  
  (1) bank \qquad (2) library
  
  (3) department store \qquad (4) mall
  
  (5) new york &
  1 \\
  \midrule
  Knights\&Knaves &
  A very special island is inhabited only by knights and knaves. Knights always tell the truth, and knaves always lie. You meet 2 inhabitants: Zoey, and Oliver. Zoey remarked, ``Oliver is not a knight''. Oliver stated, ``Oliver is a knight if and only if Zoey is a knave''. So who is Zoey?
  Choices: 
  
  (1) knight
  
  (2) knave & 2 \\
  \bottomrule
  \end{tabularx}
  \label{tab:dataset_examples}
\end{table*}
  \begin{table}[!h]
  \centering
  \small
  \caption{Shared decoding settings used in experiments.}
  \begin{tabular}{l l}
  \toprule
  Setting & Value \\
  \midrule
  Minimum tokens per step & 10 \\
  Maximum tokens per step & 2048 \\
  Maximum reasoning steps per trajectory & 50 \\
  Total maximum tokens & 4096 \\
  Temperature  & 1.0 \\
  Top-$p$, top-$k$ & 1.0, 1000 \\
  Repetition penalty & 1.2 \\
  Random seeds & $\{0,1,2,3,4\}$ \\
  \bottomrule
  \end{tabular}
  \label{tab:shared_decoding_settings}
\end{table}
  \begin{table}[!h]
  \centering
  \small
  \caption{Method-specific inference budgets used in the main experiments.}
  \begin{tabular}{l l}
  \toprule
  Method & Inference budget \\
  \midrule
  SC & $N \in \{5,25\}$ reasoning chains \\
  WSC & $N \in \{5,25\}$ reasoning chains \\
  DLE & $N=25$ reasoning chains \\
  TSMC & {$N=25$ particles} \\
  BS & \multirowcell{$N=5$ chains, $C=5$ candidate expansions} \\
  BS+SC & \multirowcell{$N=5$ chains, $C=5$ candidate expansions} \\
  VGSC & \multirowcell{$N=5$ chains, $C=5$ candidate expansions} \\
  \bottomrule
  \end{tabular}
  \label{tab:method_budgets}
\end{table}

  We evaluate on five benchmarks covering open-ended mathematical reasoning and multiple-choice scientific, commonsense, and logical reasoning. MATH and OlympiadBench are evaluated as open-ended generation tasks, while filtered text-only subsets of ScienceQA, CommonsenseQA, and Knights\&Knaves are evaluated as multiple-choice tasks. Table~\ref{tab:dataset_stats} reports subset sizes and mean reasoning-step counts for the Qwen2.5-Math-1.5B-Instruct generator.

  For multiple-choice datasets, we map each gold answer to its answer-choice index and evaluate the predicted boxed integer. Table~\ref{tab:prompt_templates} summarizes the prompts. We keep prompt semantics fixed across generator families, changing only model-specific chat-template packaging when needed.

  Table~\ref{tab:dataset_examples} provides example questions from each benchmark.

\subsection{Answer Extraction and Evaluation}
\label{app:answer_extraction_evaluation}

  We use the default answer-extraction and evaluation logic implemented in the ThinkBooster library \citep{thinkbooster2026}. All reported results are exact-match accuracies computed on the extracted final answer.

\paragraph{Answer extraction.}
  For each model output, we first extract the contents of the last complete \texttt{\textbackslash boxed\{\}} expression. If no complete boxed answer is found, we fall back to the last parseable numeric span (integer or decimal) in the generated text. For the multiple-choice datasets (ScienceQA, CommonsenseQA, and Knights\&Knaves), we explicitly prompt the model to output the answer-choice index in boxed form, e.g.\ \texttt{\textbackslash boxed\{2\}}. This allows us to use a single extraction pipeline across both open-ended and multiple-choice benchmarks.

\paragraph{Answer evaluation.}
  For ScienceQA, CommonsenseQA, and Knights\&Knaves, exact-match is computed against the mapped gold answer-choice index. For MATH and OlympiadBench, we use a math-aware exact-match evaluator. The evaluator first normalizes the extracted answer and then applies a cascade of checks: normalized string match, numeric comparison, and symbolic comparison when both the prediction and the reference can be parsed as mathematical expressions. The symbolic comparison uses SymPy parsing utilities, so algebraically equivalent expressions are counted as correct \citep{sympy}.

\subsection{Step Splitting and Decoding Details}
\label{app:step_splitting_decoding}

  Our decoding strategies operate over reasoning steps rather than individual tokens, following the step-based decoding framework implemented in ThinkBooster \citep{thinkbooster2026}.

\paragraph{Rule-based step splitting.}
  We use a rule-based detector to identify candidate step boundaries from discourse and structural markers such as \emph{first}, \emph{next}, \emph{then}, \emph{finally}, \emph{therefore}, \emph{thus}, \emph{let's check}, \emph{consider}, paragraph breaks, and numbered lists. We additionally include a small set of custom markers, including \emph{for example}, \emph{given that}, and \emph{similarly}, to improve consistency across model families.

\paragraph{Decoding settings.}
  Unless otherwise noted, all main-paper experiments use the settings in Tables~\ref{tab:shared_decoding_settings} and~\ref{tab:method_budgets}. We enforce a minimum of 10 tokens per step, cap each step at 2048 tokens, and cap each trajectory at 50 reasoning steps.

\subsection{Verifier Training Details}
\label{app:verifier_training}
  We train two verifier families: a PRM-based verifier and a ReProbe verifier \citep{ni2026efficienttesttimescalingmultistep}. For each family, we train separate models for step correctness and step continuation.

\begin{table*}[!h]
  \centering
  \small
  \setlength{\tabcolsep}{6pt}
  \caption{Hyperparameters for the two verifier families. For PRM, the step verifier is frozen and the future verifier uses last-layer QLoRA; ReProbe variants are trained separately for each supervision target.}
  \begin{tabularx}{\textwidth}{l X X}
  \toprule
  Parameter & PRM verifier & ReProbe verifier \\
  \midrule
  Architecture details
  & LoRA fine-tuning of \qwenprm; last {$4$} layers only; rank {$128$}; alpha {$256$}; dropout $0.05$; attentions and MLP projections
  & Last hidden layer from LLM as features; ReProbe uses a 2-layer transformer head with dim. 768, 8 heads, and dropout $0.1$ \\
  Trainable parameters
  & {46.1M} & 15m - 19m depending on base LLM \\
  Maximum length
  & {2048}
  & 2048 \\
  Batch size
  & {32 (1 per device, 32 accumulation steps)}
  & 16 \\
  Learning rate
  & {$10^{-4}$}
  & $10^{-4}$ \\
  Epochs
  & {5}
  & 6 \\
  Validation split
  & {$10\%$}
  & $10\%$ \\
  Training seed
  & {42}
  & 42 \\
  Regularization
  & ---
  & Weight decay $0.1$, max grad norm $1.0$ \\
  \bottomrule
  \end{tabularx}
  \label{tab:verifier_hparams}
\end{table*}
\begin{table*}[!h]
\centering
\small
\caption{Estimated inference compute. FLOPs are calculated using equation~\eqref{eq:compute_formula}. 
PRM VGSC counts a shared frozen prefix and two last-layer branches (equal-layer estimate); ReProbe VGSC counts two $19$M probes.}
\resizebox{\textwidth}{!}{
\begin{tabular}{l|rrr|rrr|rrr}
\toprule
Generator +  & \multicolumn{3}{c|}{\qwen +} & \multicolumn{3}{c|}{\llama +} & \multicolumn{3}{c}{\qwen +} \\
Verifier & \multicolumn{3}{c|}{\qwenprm} & \multicolumn{3}{c|}{\qwenprm} & \multicolumn{3}{c}{\reprobe} \\
\midrule
Method & Tokens & Params & TFLOPs & Tokens & Params & TFLOPs & Tokens & Params & TFLOPs \\
\midrule
Single sample        & 587    & 1.5B   & 2.1   & 754    & 3.0B   & 5.1   & 638    & 1.5B   & 2.2 \\
SC ($N=5$)           & 2{,}928 & 1.5B   & 9.1   & 3{,}058 & 3.0B   & 18.9  & 3{,}206 & 1.5B   & 9.9 \\
SC ($N=25$)          & 14{,}778& 1.5B   & 44.6  & 15{,}288& 3.0B   & 92.3  & 16{,}021& 1.5B   & 48.4 \\
\midrule
WSC ($N=5$)          & 2{,}964 & 8.5B   & 52.0  & 3{,}029 & 10.0B  & 62.5  & 3{,}243 & 1.519B & 10.1 \\
WSC ($N=25$)         & 14{,}791& 8.5B   & 253.1 & 15{,}260& 10.0B  & 307.1 & 16{,}036& 1.519B & 49.0 \\
\midrule
BS-Step              & 19{,}837& 8.5B   & 338.9 & 42{,}567& 10.0B  & 853.3 & 21{,}905& 1.519B & 66.8 \\
BS-Future            & 12{,}337 & 8.5B & 211.4 & 15{,}646 & 10.0B & 314.9 & 20{,}539& 1.519B & 62.7 \\
\midrule
VGSC ($\alpha=5$)    & 13{,}839 & 9.5B & 264.8 & 33{,}158 & 11.0B & 731.6 & 11{,}225& 1.538B & 34.8 \\
VGSC ($\alpha=10$)   & 15{,}179 & 9.5B & 290.2 & 32{,}970 & 11.0B & 727.5 & 11{,}240& 1.538B & 34.9 \\
\bottomrule
\end{tabular}
}
\label{tab:inference_compute}
\end{table*}

\paragraph{Training data and supervision.}
  Following~\citep{ni2026efficienttesttimescalingmultistep}, we construct verifier training data from the PRM800k corpus \citep{lightman2023letsverifystepstep}. For each generator family, we first sample 10{,}828 questions, generate 3 reasoning trajectories per question, and split each trajectory into steps using the same rule-based splitter used at test time. From every intermediate step prefix $y_{\le t}$, we then sample 10 Monte Carlo continuations using the same generation settings as at test time.

  An LLM judge \gpt annotates each trajectory with the index $k$ of its first incorrect step. We derive two supervision targets from these annotations. For the continuation target, the label for prefix $y_{\le t}$ is the fraction of its sampled continuations that reach the correct final answer:
  \begin{equation*}
    c_t = \frac{1}{10}\sum_{i=1}^{10}
    \mathds{1}\left\{
      \renewcommand{\arraystretch}{0.7}
      \begin{array}{c}
        \small \text{continuation } i \text{ reaches}\\
        \small \text{the correct final answer}
      \end{array}
    \right\}.
  \end{equation*}
  For the correctness target, if the first incorrect step is $k$, then steps $1,\dots,k-1$ receive label $1$, step $k$ receives label $0$, and all later steps are ignored.

\paragraph{Training settings.}
  For the PRM family, we use the original \qwenprm for the local correctness estimate $V$, and for continuation values $Q$, we fine-tune LoRA adapters only in its last \pending{$4$} layers (rank \pending{$128$}, scale \pending{$256$}), freezing the embedding and first \pending{$24$} layers. For the \reprobe{} family, we train the local-correctness and continuation probes separately from scratch, using the same overall training setup and implementations from~\citet{shelmanov-etal-2025-head}. The number of training epochs was selected based on the best validation-set accuracy. Table~\ref{tab:verifier_hparams} summarizes the verifier training hyperparameters.

\begin{table*}[h]
  \centering
  \small
  \caption{Accuracy $\uparrow$ across datasets and average generated tokens per sample for \llama generator with \qwenprm verifier. Results are averaged over $5$ random seeds. The best result in each column is \textbf{bold} and the second-best result is \underline{underlined}.}
  \resizebox{0.9\textwidth}{!}{
  \begin{tabular}{l|ccccc|ccc}
    \toprule
    Method & MATH & \multirowcell{Olympiad\\Bench} & SciQA & CsQA & K\&K & \multirowcell{Average\\Accuracy} & \multirowcell{Average\\Tokens}  & \multirowcell{Estimated\\TFLOPs} \\
\midrule
 \rowcolor{gray!20}
 \multicolumn{9}{c}{\textit{No-verifier baselines}} \\
\midrule
Single sample & 36.2\pmstd{0.5} & 11.2\pmstd{0.3} & 51.7\pmstd{0.6} & 42.3\pmstd{0.3} & 29.0\pmstd{1.9} & 34.1\pmstd{0.4} & 754 & 5.1 \\
SC $N=5$ & 38.6\pmstd{0.9} & 10.0\pmstd{0.2} & 69.5\pmstd{0.7} & 59.8\pmstd{0.2} & 36.6\pmstd{2.0} & 42.9\pmstd{0.5} & 3{,}058 & 18.9 \\
SC $N=25$ & 50.9\pmstd{0.1} & 16.4\pmstd{0.3} & 82.7\pmstd{0.5} & 71.4\pmstd{0.2} & 49.4\pmstd{1.7} & 54.2\pmstd{0.4} & 15{,}288 & 92.3 \\
\midrule
 \rowcolor{gray!20}
 \multicolumn{9}{c}{\textit{Post-hoc verifier aggregation}} \\
\midrule
WSC $N=5$ & 46.4\pmstd{0.5} & 13.7\pmstd{0.3} & 68.5\pmstd{0.7} & 61.1\pmstd{0.6} & 34.8\pmstd{1.3} & 44.9\pmstd{0.3} & 3{,}029 & 62.5 \\
WSC $N=25$ & 57.3\pmstd{0.5} & 19.6\pmstd{0.2} & 80.7\pmstd{0.5} & 71.8\pmstd{0.5} & 47.0\pmstd{2.2} & 55.3\pmstd{0.5} & 15{,}260 & 307.1 \\
DLE & 59.8\pmstd{0.4} & 28.3\pmstd{0.2} & 77.6\pmstd{0.1} & \underline{72.2}\pmstd{0.2} & {58.0}\pmstd{1.0} & 59.2\pmstd{0.2} & 14{,}953 & 301.0 \\
\midrule
 \rowcolor{gray!20}
 \multicolumn{9}{c}{\textit{Verifier-guided search baselines}} \\
\midrule
BS-Step & 52.3\pmstd{0.5} & 16.4\pmstd{0.3} & 53.0\pmstd{0.5} & 43.7\pmstd{0.6} & 14.0\pmstd{1.6} & 35.9\pmstd{0.4} & 42{,}567 & 853.3 \\
BS-Future & 64.8\pmstd{0.4}& 27.3\pmstd{0.4}& 76.1\pmstd{0.8}& 67.0\pmstd{0.4}& 41.0\pmstd{1.0}& 55.2\pmstd{0.3}& 15{,}646 & 314.9 \\
BS+SC-Step & 61.0\pmstd{0.3} & 22.3\pmstd{0.2} & 78.1\pmstd{0.5} & 65.6\pmstd{0.2} & 54.4\pmstd{1.8} & 56.3\pmstd{0.4} & 42{,}944 & 860.8 \\
BS+SC-Future & 65.7\pmstd{0.5}& 28.1\pmstd{0.4}& 79.3\pmstd{0.3}& 70.7\pmstd{0.6}& 61.5\pmstd{2.2}& 61.1\pmstd{0.5}& 15{,}270 & 307.3 \\
TSMC & 57.6\pmstd{0.2}& 21.4\pmstd{0.4}& 80.7\pmstd{0.4}& 69.0\pmstd{0.6}& 54.0\pmstd{1.8}& 56.5\pmstd{0.4}& 15{,}873 & 351.3 \\
\midrule
 \rowcolor{gray!20}
 \multicolumn{9}{c}{\textit{Verifier-guided self-consistency (ours)}} \\
\midrule
VGSC $\alpha=1$ & 63.6\pmstd{0.6}& 24.5\pmstd{0.3}& 83.7\pmstd{0.4}& 71.5\pmstd{0.2}& 60.5\pmstd{1.8}& 60.8\pmstd{0.4}& 28{,}715 & 633.9 \\
VGSC $\alpha=5$ & \underline{67.4}\pmstd{0.4}& \underline{28.7}\pmstd{0.2}& \underline{84.0}\pmstd{0.4}& \underline{72.2}\pmstd{0.3}& \underline{62.5}\pmstd{2.1}& \underline{63.0}\pmstd{0.4}& 33{,}158 & 731.6 \\
VGSC $\alpha=10$ & 67.0\pmstd{0.7}& \textbf{30.0}\pmstd{0.2}& 83.4\pmstd{0.4}& \textbf{72.5}\pmstd{0.5}& 59.5\pmstd{1.3}& 62.5\pmstd{0.3}& 32{,}970 & 727.5 \\
VGSC $\alpha=20$ & \textbf{68.6}\pmstd{0.2}& 28.3\pmstd{0.3}& \textbf{84.4}\pmstd{0.3}& 71.1\pmstd{0.3}& \textbf{64.0}\pmstd{2.9}& \textbf{63.3}\pmstd{0.6}& 33{,}340 & 735.6 \\
    \bottomrule
  \end{tabular}
  }
  \label{tab:llama_prm}
\end{table*}

\section{Inference Compute Estimate}
\label{app:inference_compute}

  We estimate inference compute as model FLOPs. For a dense decoder-only Transformer, we approximate a forward pass over $T$ tokens with a $P$-parameter model as $2PT$ FLOPs \citep{Kaplan2020ScalingLF}. Let $T_{\mathrm{gen}}$ be the average generated-token count, $T_{\mathrm{prompt}}$ is the mean prompt length, and $P_m$ the parameter count of counted component $m$ (generator or verifier). With cached incremental decoding, we estimate

  \begin{equation}
    \widehat C = 2(T_{\mathrm{gen}} + T_{\mathrm{prompt}})\sum_m P_m.
    \label{eq:compute_formula}
  \end{equation}

  Table~\ref{tab:inference_compute} reports a shared-prefix inference estimate, not measured compute of our separate-PRM implementation. For a {$28$}-layer PRM with only the last {$4$} layers adapted, two verifier branches count {$8$B} effective parameters under the equal-layer approximation, instead of {$14$B}. Including the generator gives {$9.5$B} for Qwen and {$11$B} for Llama. A single future PRM still requires the full {$7$B} forward pass. Savings require identical input tokens and cached-prefix states for both branches; they do not follow from freezing weights alone.

  When the verifier is lightweight relative to the generator, VGSC's generated-token savings can outweigh the cost of the additional future-outcome verifier. This is reflected in the \reprobe setting: counting one $19$M probe for WSC/BS and two $19$M probes for VGSC, VGSC has lower estimated FLOPs than the high-budget \reprobe baselines while using substantially fewer generated tokens. We view this lightweight-verifier regime as natural in practical deployments, where verifiers are often chosen to be smaller and cheaper than the generator they guide.

\section{Additional Results}
\label{app:additional_results}
  The main text reports results for the \qwen model using the \qwenprm verifier. Here, we present additional results for \llama and \gemma with the same verifier. For \qwen, we also evaluate a second verifier family, \reprobe~\citep{ni2026efficienttesttimescalingmultistep}. Tables~\ref{tab:llama_prm}, \ref{tab:gemma_prm} and~\ref{tab:qwen_reprobe} show that the main trend is consistent across LLMs and verifiers: VGSC remains competitive and typically outperforms the corresponding baselines, suggesting that the benefits of verifier-guided self-consistency are not limited to a single LLM or verifier family.

\begin{table*}[!ht]
  \centering
  \small
  \caption{Accuracy $\uparrow$ across datasets and average generated tokens per sample for \gemma generator with \qwenprm verifier. Results are averaged over $5$ random seeds. The best result in each column is \textbf{bold} and the second-best result is \underline{underlined}.}
  \resizebox{0.9\textwidth}{!}{
  \begin{tabular}{l|ccccc|ccc}
    \toprule
    Method & MATH & \multirowcell{Olympiad\\Bench} & SciQA & CsQA & K\&K & \multirowcell{Average\\Accuracy} & \multirowcell{Average\\Tokens}  & \multirowcell{Estimated\\TFLOPs} \\
\midrule
 \rowcolor{gray!20}
 \multicolumn{9}{c}{\textit{No-verifier baselines}} \\
\midrule
Single sample & 67.2\pmstd{0.5} & 29.6\pmstd{0.3} & 68.9\pmstd{0.5} & 68.1\pmstd{0.7} & 60.0\pmstd{1.6} & 58.8\pmstd{0.4} & 845 & 7.5 \\
SC $N=5$ & 74.7\pmstd{0.4} & 36.7\pmstd{0.2} & 75.1\pmstd{0.7} & 70.4\pmstd{0.4} & 66.8\pmstd{1.0} & 64.7\pmstd{0.3} & 4{,}315 & 35.3 \\
SC $N=25$ & 78.6\pmstd{0.4} & 41.6\pmstd{0.2} & 74.3\pmstd{0.3} & 72.9\pmstd{0.3} & 67.6\pmstd{1.1} & 67.0\pmstd{0.2} & 21{,}581 & 173.4 \\
\midrule
 \rowcolor{gray!20}
 \multicolumn{9}{c}{\textit{Post-hoc verifier aggregation}} \\
\midrule
WSC $N=5$ & 76.0\pmstd{0.4} & 37.9\pmstd{0.4} & 79.3\pmstd{0.4} & 72.8\pmstd{0.3} & 68.4\pmstd{1.0} & 66.9\pmstd{0.3} & 4{,}308 & 96.9 \\
WSC $N=25$ & 80.8\pmstd{0.3} & \underline{43.4}\pmstd{0.3} & 80.8\pmstd{0.3} & \textbf{74.8}\pmstd{0.2} & 71.4\pmstd{1.1} & 70.2\pmstd{0.2} & 21{,}573 & 476.7 \\
DLE & 77.4\pmstd{0.5} & 42.0\pmstd{0.4} & 80.6\pmstd{0.4} & \underline{74.4}\pmstd{0.3} & 65.2\pmstd{1.7} & 67.9\pmstd{0.3} & 13{,}285 & 294.4 \\
\midrule
 \rowcolor{gray!20}
 \multicolumn{9}{c}{\textit{Verifier-guided search baselines}} \\
\midrule
BS-Step & 73.3\pmstd{0.4} & 33.4\pmstd{0.2} & 80.6\pmstd{0.5} & 70.7\pmstd{0.4} & 70.4\pmstd{0.9} & 65.7\pmstd{0.2} & 29{,}586 & 653.0 \\
BS-Future & 77.6\pmstd{0.3}& 37.7\pmstd{0.3}& 81.3\pmstd{0.4}& 72.9\pmstd{0.3}& 65.5\pmstd{0.9}& 67.0\pmstd{0.2}& 11{,}329 & 251.4 \\
BS+SC-Step & 75.6\pmstd{0.4} & 36.9\pmstd{0.2} & 79.6\pmstd{0.5} & 71.2\pmstd{0.5} & \textbf{73.5}\pmstd{1.2} & 67.4\pmstd{0.3} & 29{,}562 & 652.5 \\
BS+SC-Future & 76.9\pmstd{0.2}& 40.3\pmstd{0.3}& 79.0\pmstd{0.2}& 70.8\pmstd{0.3}& 67.0\pmstd{0.8}& 66.8\pmstd{0.2}& 11{,}562 & 256.5 \\
TSMC & 79.5\pmstd{0.3}& 42.3\pmstd{0.3}& 81.1\pmstd{0.6}& {71.6}\pmstd{0.2}& 71.0\pmstd{1.3}& 69.1\pmstd{0.3}& 16{,}295 & 393.4 \\
\midrule
 \rowcolor{gray!20}
 \multicolumn{9}{c}{\textit{Verifier-guided self-consistency (ours)}} \\
\midrule
VGSC $\alpha=1$ & 81.3\pmstd{0.4}& 42.9\pmstd{0.5}& \underline{82.0}\pmstd{0.4}& 73.2\pmstd{0.3}& 72.0\pmstd{0.7}& \underline{70.3}\pmstd{0.2}& 18{,}846 & 454.6 \\
VGSC $\alpha=5$ & \underline{81.5}\pmstd{0.3}& \textbf{43.6}\pmstd{0.5}& \textbf{82.3}\pmstd{0.4}& 73.4\pmstd{0.3}& 70.5\pmstd{0.9}& \underline{70.3}\pmstd{0.2}& 19{,}553 & 471.6 \\
VGSC $\alpha=10$ & \textbf{82.0}\pmstd{0.2}& 43.1\pmstd{0.2}& 80.6\pmstd{0.3}& 73.4\pmstd{0.3}& 69.5\pmstd{1.2}& 69.7\pmstd{0.3}& 19{,}740 & 476.1 \\
VGSC $\alpha=20$ & 81.0\pmstd{0.3}& 43.0\pmstd{0.1}& 81.0\pmstd{0.3}& 73.6\pmstd{0.2}& \textbf{73.5}\pmstd{1.0}& \textbf{70.4}\pmstd{0.2}& 19{,}696 & 475.0 \\
    \bottomrule
  \end{tabular}
  }
  \label{tab:gemma_prm}
\end{table*}
\begin{table*}[!ht]
  \centering
  \small
  
  \caption{Accuracy $\uparrow$ across datasets and average generated tokens per sample for \qwen generator with \reprobe verifier. Results are averaged over $5$ random seeds. The best result in each column is \textbf{bold} and the second-best result is \underline{underlined}.}
  \resizebox{0.9\textwidth}{!}{
  \begin{tabular}{l|ccccc|cc}
    \toprule
    Method & MATH & \multirowcell{Olympiad\\Bench} & SciQA & CsQA & K\&K & \multirowcell{Average\\Accuracy} & \multirowcell{Average\\Tokens} \\
\midrule
 \rowcolor{gray!20}
 \multicolumn{8}{c}{\textit{No-verifier baselines}} \\
\midrule
Single sample & 66.8\pmstd{0.4} & 29.9\pmstd{0.4} & 46.2\pmstd{0.4} & 29.8\pmstd{0.4} & 38.0\pmstd{1.0} & 42.1\pmstd{0.2} & 638 \\
SC (\(N=5\)) & 70.0\pmstd{0.4} & 34.2\pmstd{0.4} & 57.0\pmstd{0.4} & 39.8\pmstd{0.4} & 53.0\pmstd{1.0} & 50.8\pmstd{0.2} & 3{,}206 \\
SC (\(N=25\)) & 73.6\pmstd{0.4} & 38.5\pmstd{0.4} & \underline{63.8}\pmstd{0.4} & \textbf{47.8}\pmstd{0.4} & 54.0\pmstd{1.0} & \underline{55.5}\pmstd{0.2} & 16{,}021 \\
\midrule
 \rowcolor{gray!20}
 \multicolumn{8}{c}{\textit{Post-hoc verifier aggregation}} \\
\midrule
WSC (\(N=5\)) & 72.7\pmstd{0.4} & 35.3\pmstd{0.4} & 56.8\pmstd{0.4} & 38.0\pmstd{0.4} & 51.2\pmstd{1.0} & 50.8\pmstd{0.2} & 3{,}243 \\
WSC (\(N=25\)) & \textbf{75.0}\pmstd{0.4} & \underline{40.0}\pmstd{0.4} & 62.2\pmstd{0.4} & 43.6\pmstd{0.4} & 54.2\pmstd{1.0} & 55.0\pmstd{0.2} & 16{,}036 \\
DLE & 74.4\pmstd{0.5} & 39.3\pmstd{0.5} & 62.0\pmstd{0.3} & 44.4\pmstd{0.5} & 55.6\pmstd{2.0} & 55.1\pmstd{0.4} & 13{,}510 \\
\midrule
 \rowcolor{gray!20}
 \multicolumn{8}{c}{\textit{Verifier-guided search baselines}} \\
\midrule
BS-Step & 73.2\pmstd{0.4} & 35.1\pmstd{0.4} & 33.7\pmstd{0.4} & 19.2\pmstd{0.4} & 32.0\pmstd{1.0} & 38.6\pmstd{0.2} & 21{,}905 \\
BS-Future & 72.0\pmstd{0.4} & 34.5\pmstd{0.4} & 39.8\pmstd{0.4} & 25.7\pmstd{0.4} & 41.2\pmstd{1.0} & 42.6\pmstd{0.2} & 20{,}539 \\
BS+SC-Step & 71.7\pmstd{0.4} & 36.2\pmstd{0.4} & 56.7\pmstd{0.4} & 38.2\pmstd{0.4} & 54.2\pmstd{1.0} & 51.4\pmstd{0.2} & 21{,}891 \\
BS+SC-Future & 73.5\pmstd{0.4} & 37.5\pmstd{0.4} & 55.7\pmstd{0.4} & 38.0\pmstd{0.4} & 56.6\pmstd{1.0} & 52.3\pmstd{0.2} & 20{,}551 \\
TSMC & 68.8\pmstd{0.4} & 33.8\pmstd{0.4} & 56.8\pmstd{0.4} & 38.0\pmstd{0.4} & 50.0\pmstd{1.0} & 49.5\pmstd{0.2} & 12{,}827 \\
\midrule
 \rowcolor{gray!20}
 \multicolumn{8}{c}{\textit{Verifier-guided self-consistency (ours)}} \\
\midrule
VGSC (\(\alpha=1\)) & 72.8\pmstd{0.4} & 38.8\pmstd{0.4} & 60.7\pmstd{0.4} & 39.2\pmstd{0.4} & 57.0\pmstd{1.0} & 53.7\pmstd{0.2} & 10{,}917 \\
VGSC (\(\alpha=5\)) & 72.9\pmstd{0.4} & 39.6\pmstd{0.4} & 62.7\pmstd{0.4} & 41.1\pmstd{0.4} & \underline{58.8}\pmstd{1.0} & 55.0\pmstd{0.2} & 11{,}225 \\
VGSC (\(\alpha=10\)) & 73.7\pmstd{0.4} & 38.4\pmstd{0.4} & \textbf{64.0}\pmstd{0.4} & 41.7\pmstd{0.4} & 56.8\pmstd{1.0} & 54.9\pmstd{0.2} & 11{,}240 \\
VGSC (\(\alpha=20\)) & \underline{74.8}\pmstd{0.4} & \textbf{40.6}\pmstd{0.4} & 60.7\pmstd{0.4} & \underline{45.1}\pmstd{0.4} & \textbf{59.6}\pmstd{1.0} & \textbf{56.2}\pmstd{0.2} & 11{,}680 \\
    \bottomrule
  \end{tabular}
  }
  \label{tab:qwen_reprobe}
\end{table*}

\clearpage

\section{Baselines Description}
\label{app:baselines}
  This appendix briefly describes the baseline decoding strategies used in our experiments. Throughout, $\xv$ denotes the input problem, $\yv = (y_1, \dots, y_T)$ a reasoning trajectory consistent of $T$ reasoning steps, $P(\yv \mid \xv)$ the base generator distribution, and $A(\yv)$ the extracted final answer.

\subsection{No-verifier baselines}

\paragraph{Single sample.}
  Single sample generates one reasoning trajectory
  \begin{equation*}
    \yv \sim P(\cdot \mid \xv)
  \end{equation*}
  and returns the extracted final answer $A(\yv)$. This is the minimal-compute baseline and can be viewed as the special case of self-consistency with $N=1$.

\paragraph{Self-consistency (SC).}
  Self-consistency \citep{wang2023selfconsistencyimproveschainthought} samples $N$ independent reasoning trajectories
  \begin{equation*}
    \yv^1, \dots, \yv^N \overset{i.i.d.}{\sim} P(\cdot \mid \xv)
  \end{equation*}
  and predicts the majority answer:
  \begin{equation}
    \mathrm{SC}(\xv)
    =
    \arg\max_a \sum_{i=1}^N \mathds{1}\left( A(\yv^i) = a \right).
  \end{equation}
  SC does not use verifier information during generation or aggregation.

\subsection{Post-hoc verifier aggregation}

\paragraph{Weighted self-consistency (WSC).}
  Weighted self-consistency \citep{kang2025scalablebestofnselectionlarge} uses the same sampled trajectories as SC, but weights each vote by a verifier score for the full trajectory. Concretely, it predicts
  \begin{align}
    &\mathrm{WSC}(\xv) = \\
    &\arg\max_a \sum_{i=1}^N V(\yv^i \mid \xv)\, \mathds{1}\left( A(\yv^i) = a \right). \nonumber
  \end{align}
  Thus, WSC uses verifier information only after complete reasoning trajectories have already been generated, without affecting the decoding process itself.

\textbf{DLE.} Distinct Leaf Enumeration (DLE; \citet{li_efficient_2026}) is a deterministic test-time procedure that explores a truncated next-token decoding tree based on sampling rules and returns up to $k$ distinct completed reasoning chains. In our experiments, DLE is used strictly as a post-hoc verifier baseline: the base generator runs the external DLE enumerator (leaf budget $k=25$, temperature $=1$, top-$p=1$, truncation parameter $\varepsilon=0.05$). No verifier signal is used during tree expansion. Concretely, let $q(\cdot \mid \mathbf{x}_{<t})$ denote the truncated next-token distribution with active set $A(\mathbf{x}_{<t}) = \{v \in \mathcal{V} : p_t(v \mid \mathbf{x}_{<t}) > \varepsilon\}$, and let $Q(\mathbf{x}_{1:T}) = \prod_{t=1}^{T} q(x_t \mid \mathbf{x}_{<t})$ be the induced sequence probability. Starting from a greedy rollout, DLE iteratively selects the unexplored branch $\mathbf{x}_{<t} \circ v$ with the largest path probability and completes it greedily,
\begin{equation}
    (\mathbf{x}_{<t}^\star, v^\star) = \arg\max_{(\mathbf{x}_{<t}, v) \in \mathcal{F}} Q(\mathbf{x}_{<t} \circ v),
\end{equation}
where $\mathcal{F}$ denotes the frontier of unexplored branch points in the truncated tree, until $k$ distinct leaves $\mathbf{y}^1, \ldots, \mathbf{y}^k$ are produced. After decoding, each leaf is split into reasoning steps with the same rule-based step detector used elsewhere in this work, and the PRM assigns a correctness score to every step. Per-leaf verifier weights are obtained by aggregating the product of these step scores. Equivalent extracted answers are clustered, and the final prediction is obtained by WSC:
\begin{equation}
    \mathrm{DLE}(\mathbf{x}) = \arg\max_a \sum_{i=1}^{k} V(\mathbf{y}^i \mid \mathbf{x}) \, \mathbf{1}\!\left[\mathcal{A}(\mathbf{y}^i) = a\right].
\end{equation}

\subsection{Verifier-guided search baselines}

\paragraph{Beam search (BS).}
  Beam search maintains the top $N$ partial reasoning chains during decoding. At step $t$, suppose the current beam consists of prefixes
  \begin{equation*}
    \yv_{<t}^1, \dots, \yv_{<t}^N.
  \end{equation*}
  For each prefix $\yv_{<t}^i$, we sample $C$ candidate next steps
  \begin{equation*}
    y_t^{(i,1)}, \dots, y_t^{(i,C)}
    \overset{i.i.d.}{\sim}
    P(\cdot \mid \xv, \yv_{<t}^i),
  \end{equation*}
  forming $N \cdot C$ expanded prefixes
  \begin{equation*}
    \yv_{\le t}^{(i,j)} = \yv_{<t}^i \circ y_t^{(i,j)}.
  \end{equation*}
  These expanded prefixes are scored, ranked globally, and only the top $N$ are kept in the beam.

  We consider two alternative scoring rules. In \textit{BS-Step}, we score a prefix by its accumulated step-level verifier score,
  \begin{equation}
    S_{\mathrm{step}}(\yv_{\le t}, \xv)
    :=
    \prod_{s=1}^{t}
    V(y_s \mid \xv, \yv_{<s}),
  \end{equation}
  i.e., using $V$ as the beam score. In \textit{BS-Future}, we score a prefix using the future-outcome predictor,
  \begin{equation}
    S_{\mathrm{future}}(\yv_{\le t}, \xv)
    :=
    \widehat{Q}(\yv_{\le t}, \xv),
  \end{equation}
  i.e., using $Q$ as the beam score. Decoding continues until the beam trajectories are completed, and the final prediction is the answer extracted from the top completed trajectory under the corresponding scoring rule.

\paragraph{Beam search with self-consistency (BS+SC).}
  Beam search with self-consistency uses the same beam expansion and prefix scoring procedure as BS, but aggregates the final answers of the completed beam trajectories by majority vote instead of returning only the top one. If $\yv^1, \dots, \yv^N$ denote the final completed beam trajectories, then
  \begin{equation}
    \mathrm{BSSC}(\xv)
    =
    \arg\max_a
    \sum_{i=1}^N
    \mathds{1}\left( A(\yv^i) = a \right).
  \end{equation}
  As with BS, we evaluate two variants: \textit{BS+SC-Step}, which uses $S_{\mathrm{step}}$, and \textit{BS+SC-Future}, which uses $S_{\mathrm{future}}$ during beam construction.

\paragraph{TSMC.}
  Twisted Sequential Monte Carlo (TSMC; \citealp{zhao2024probabilisticinferencelanguagemodels}) performs verifier-guided decoding by maintaining $N$ particles, where each particle is a partial reasoning trajectory. At step $t$, each unfinished particle $\yv_{<t}^i$ is extended by sampling one next reasoning step from the base generator,
  \begin{equation*}
    y_t^i \sim P(\cdot \mid \xv, \yv_{<t}^i),
    \qquad
    \yv_{\le t}^i = \yv_{<t}^i \circ y_t^i .
  \end{equation*}
  During decoding, TSMC uses the future-outcome predictor
  \begin{equation}
    \widehat{Q}(\yv_{\le t}, \xv)
    \approx
    \mathbb{E}_{\yv_{>t} \sim P(\cdot \mid \xv, \yv_{\le t})}
    \left[ V(\yv \mid \xv) \right]
  \end{equation}
  to score partial trajectories and compute non-terminal importance weights
  \begin{equation}
    w_t^i
    =
    \frac{\widehat{Q}(\yv_{\le t}^i, \xv)}{\widehat{Q}(\yv_{<t}^i, \xv)}.
  \end{equation}
  Unfinished particles are then resampled proportionally to $w_t^i$, so computation is concentrated on prefixes predicted to lead to strong final solutions. Once a particle terminates, the terminal verifier $V$ is used only at the end through
  \begin{equation}
    w_{T_i}^i
    =
    \frac{V(\yv^i \mid \xv)}{\widehat{Q}(\yv_{<T_i}^i, \xv)}.
  \end{equation}
  The final prediction is obtained by weighted answer aggregation over completed particles:
  \begin{equation}
    \mathrm{TSMC}(\xv)
    =
    \arg\max_a
    \sum_{i=1}^N
    w_{T_i}^i \,
    \mathds{1}\left( A(\yv^i)=a \right).
  \end{equation}

  Compared with the terminal-potential TSMC variant we evaluate, VGSC uses both local and long-horizon verifier information during decoding. TSMC guides intermediate search through $Q$ and applies the terminal verifier $V$ only at completion, whereas VGSC also incorporates $V$ during intermediate decoding. This may improve robustness when local and future verifier signals capture complementary failure modes.

\section{Computational Budget}
\label{app:comp_budget}
  For the main experiments, we evaluated 15 methods on 5 datasets using 5 random seeds across 4 model configurations (3 LLMs with PRM and 1 LLM with \reprobe{}). Each run required approximately 3 hours on 2 NVIDIA RTX 4090 GPUs (24\,GB each). This results in a total computational cost of approximately $9{,}000$ GPU-hours. The ablation studies required a comparable amount of computation, bringing the overall budget to approximately 18{,}000 GPU-hours.

\section{Use of Large Language Models}
\label{app:llm_usage}
  This work investigates large language models as the main object of study. In addition, LLM-based tools, such as ChatGPT, were used in a limited and auxiliary capacity to assist with tasks including language polishing, grammar checking, and code debugging. All such uses were subject to human oversight and verification.

\end{document}